\documentclass{article}

\PassOptionsToPackage{numbers, compress}{natbib}

\usepackage[preprint]{neurips_2024}

\usepackage[utf8]{inputenc} 
\usepackage[T1]{fontenc}    
\usepackage{hyperref}       
\usepackage{url}            
\usepackage{booktabs}       
\usepackage{amsfonts}       
\usepackage{nicefrac}       
\usepackage{microtype}      
\usepackage{xcolor}         

\usepackage{amsmath}
\usepackage{amssymb}
\usepackage{mathtools}
\usepackage{amsthm}
\usepackage{subcaption}

\usepackage{wrapfig}
\usepackage{graphicx}
\usepackage{hyperref}

\theoremstyle{plain}

\theoremstyle{definition}

\theoremstyle{remark}

\title{The Unreasonable Effectiveness of Solving Inverse Problems with Neural Networks}

\author{%
  Philipp Holl\\
  Technical University of Munich \\
  \texttt{philipp.holl@tum.de} \\
  \And 
  Nils Thuerey\\
  Technical University of Munich \\
  \texttt{nils.thuerey@tum.de} \\
}

\usepackage{mdframed}
\usepackage{thmtools}
\newmdtheoremenv[
  backgroundcolor=gray!10,
  linecolor=black,
  linewidth=1pt,
  topline=false,
  bottomline=false,
  rightline=false,
  font={\bfseries}
]{boxedtheorem}{Theorem}
\newmdtheoremenv[
  backgroundcolor=gray!10,
  linecolor=black,
  linewidth=1pt,
  topline=false,
  bottomline=false,
  rightline=false,
  font={\bfseries}
]{boxedpreposition}{Preposition}
\newmdtheoremenv[
  backgroundcolor=gray!10,
  linecolor=black,
  linewidth=1pt,
  topline=false,
  bottomline=false,
  rightline=false,
  font={\bfseries}
]{boxedtheorem2}{Theorem}

\begin{document}

\maketitle

\begin{abstract}
Finding model parameters from data is an essential task in science and engineering, from weather and climate forecasts to plasma control.
Previous works have employed neural networks to greatly accelerate finding solutions to inverse problems.
Of particular interest are end-to-end models which utilize differentiable simulations in order to backpropagate feedback from the simulated process to the network weights and enable roll-out of multiple time steps.
So far, it has been assumed that, while model inference is faster than classical optimization, this comes at the cost of a decrease in solution accuracy.
We show that this is generally not true.
In fact, neural networks trained to learn solutions to inverse problems can find better solutions than classical optimizers even on their training set.
To demonstrate this, we perform both a theoretical analysis as well an extensive empirical evaluation on challenging problems involving local minima, chaos, and zero-gradient regions.
Our findings suggest an alternative use for neural networks: rather than generalizing to \emph{new} data for fast inference, they can also be used to find better solutions on \emph{known} data.
\end{abstract}

\newcommand{\myvspace}{\vspace{-0.2cm}}

\section{Introduction}
Estimating model parameters by solving inverse problems~\cite{tarantola2005inverse} is a central task in scientific research, from 
    detecting gravitational waves~\cite{RealtimeMLLIGO} to
    controlling plasma flows~\cite{maingi2019fesreport} to
    searching for neutrinoless double-beta decay~\cite{GERDA1, MAJORANA}.
Recent work has explored the possibility of learning solutions to inverse problems directly using data-driven techniques~\cite{shmakov2023end, lucas2018using,ScaleInvariantTraining, HIG, mukherjee2021end,li2020nett}.
This is in large part motivated by speedups achieved by trained models, such as by predicting certain quantities of simulations~\cite{tompson2017, kochkov2021_Machine}, correcting simulation errors in order to run at lower resolutions~\cite{um2020solver} or outright replacing hand-written solvers~\cite{kim2019,sanchez2020learning, stachenfeld2021learned, pathak2022fourcastnet}.
%
%
Given a large enough data set of related inverse problems
\begin{equation} \label{eq:inverse-problem}
    x_i^* = \textrm{arg min}_{x_i} \mathcal L(F(x_i \,|\, \gamma_i), y_i),
\end{equation}
each parameterized by a target output $y_i$ and other observed quantities $\gamma_i$, machine learning models can be trained to predict solutions $x^*_i$. Here $\mathcal L$ denotes the error metric and $F$ the differentiable forward process which can be simulated numerically.
In the absence of gradients from the simulator, it has been shown that learning the forward relationship from data can yield a suitable proxy for optimizing corresponding inverse problems~\cite{schenck2018spnets, NeuralAdjoint, allen2022physical}.

Models for solving inverse problems can be trained in two fundamentally different ways, \emph{supervised} or \emph{end-to-end}.
Supervised training requires precomputed solutions $\hat x^*$ which the network is given as labels.
While this works well for unimodal problems, the presence of multiple optima can severely hamper the network predictions, as the network then minimizes the distance to each occurring solution rather than the actual error metric, and the mean of multiple solutions is generally not a solution.
End-to-end training, on the other hand, avoids this problem by back-propagating the gradient of the error metric to the network weights, thereby minimizing the difference between target output and the output resulting from the predicted solution.
However, if the individual problems are hard to optimize, this similarly affects network training, as it relies on the same gradient feedback as classical optimizers.

Independent of the employed method, the trained network can typically infer approximate solutions much faster than competing classical optimizers like BFGS~\cite{liu1989lbfgs}, Gauss-Newton~\cite{GaussNewton} or others~\cite{NumericalRecipes}.
Previous work has assumed that this advantage in speed comes at the cost of the solution quality~\cite{phiflow}.
We challenge this assumption and show that, while machine-precision accuracy is indeed difficult to obtain with neural networks, the predicted solutions can still be much closer to a global optimum than what classical optimizers typically achieve.

In order to show this, we focus on hard-to-optimize problems, where classical optimizers struggle to find a global optimum.
This includes problems with local minima, problems with zero-gradient regions, and chaotic problems.
We find that employing neural networks can alleviate all of these difficulties, with prediction quality increasing with the number of available inverse problems.

In particular, we make the following contributions:
(i) We observe that using neural networks benefits the optimization of inverse problems in terms of quality beyond what has been reported previously.
(ii) We analyze the learning behavior to show that the benefits scale approximately with the square root of the number of examples.
(iii) We perform extensive experiments to demonstrate and investigate this behavior on toy examples as well as challenging inverse problems.
(iv) We formulate our framework in such a way that it can be used to optimize any set of inverse problems, without requiring prior knowledge. This allows our approach to be used as a drop-in component or replacement for classical optimizers. Our source code demonstrates this for the included experiments.

\paragraph{Related work.}
Inverse problems have been tacked by a large variety of methods~\cite{NumericalOptimizationBook}.
With the advent of machine learning, researchers have turned to neural networks in order to solve inverse problems in imaging~\cite{DL4InverseImagingProblems,aggarwal2018modl}, math~\cite{khalil2017learning} and physics~\cite{ScaleInvariantTraining, HIG}, among others.
Contextualized learning~\cite{lengerich2023contextualized} tries to recover the sample-specific factors and has been used in medical research~\cite{lengerich2022automated,al2018personalized}.
Neural networks have also been integrated into dual dynamic programming~\cite{dai2021neural} to solve multi-stage stochastic optimization problems.
Normalizing flows can be used to map probability distributions~\cite{kobyzev2020normalizing} and have been demonostrated to be suitable for solving inverse problems as well~\cite{wildberger2024flow}.

In the context of image reconstruction, the inductive bias of convolutional networks has been shown to benefit the optimization of individual inverse problems with image solution spaces~\cite{DeepImagePrior, NeuralReparameterizationWorkshop}.
Various architectures and methods have also been explored to solve inverse problems involving PDEs on grids, such as Fourier neural operators~\cite{li2020fourier} and diffusion models~\cite{holzschuh2023solving}.

Underlying many of these approaches are differentiable simulations, required to obtain gradients of the inverse problem.
These can be used in iterative optimization or to train neural networks.
Many recent software packages have demonstrated this use of differentiable simulations, with general frameworks~\cite{hu2019difftaichi,JaxMD,phiflow} and specialized simulators~\cite{DiffFluids2021,DiffCloth2019}. 


Physics-informed neural networks~\cite{raissi2019pinn} encode solutions to optimization problems in the network weights themselves.
They model a continuous solution to an ODE or PDE and are trained by formulating a loss function based on the differential equation, and have been explored for a variety of directions~\cite{yang2019predictive,lu2021physics,krishnapriyan2021characterizing}.
However, as these approaches rely on loss terms formulated with neural network derivatives, they do not apply to general inverse problems.

The training process of neural networks themselves can also be framed as an inverse problem, and employing learning models to aid this optimization is referred to as \emph{meta-learning}~\cite{MetaLearningReview}.
This approach has been used to extract physical parameters in multi-environment settings~\cite{blanke2024interpretable}.

\section{Joint optimization of inverse problems} \label{sec:joint-optimization}

Consider a set of $N$ inverse problems as in Eq.~\ref{eq:inverse-problem}.
This setup essentially applies to all sets of inverse problems, but we focus on similar problems, i.e. where  $x_i^*$ has a learnable dependence on $y_i$ and $\gamma_i$.
We assume that $F$ is differentiable and can be approximately simulated, but allow for hidden information and stochasticity preventing $y_i$ from being perfectly fit.
Classical optimizers may struggle to solve these problems if $F$ is chaotic, has local minima or contains zero-gradient regions.

When training a neural network to predict solutions using end-to-end training, a large number of inverse problems are optimized jointly.
Not only does this allow for fast inference of solutions on unseen data, it also has a profound impact on the found $x_i^*$ of the training set.
To see how joint optimization affects the optimization, we think of the loss landscape $L_i \equiv \mathcal L(F, y_i) = \text{signal} + \text{noise}$ as consisting of a \emph{signal} component which points toward a desirable solution, and a \emph{noise} component which introduces unwanted features like local minima, chaotic behavior, etc.
Naturally, this decomposition cannot actually be performed since the components are inherently linked and we have not assumed any prior knowledge about the functional form of $F$.
Nevertheless we can approximate the noise by its Fourier series which only requires piece-wise continuity and a finite number of extrema and discontinuities.
Then the loss can be written as
\begin{equation} \label{eq:fourier-model}
    L_i(x) = \underbrace{\lambda_i |x_i - x_i^*|}_\text{Signal} + \underbrace{\sum_{j=1}^m -A_{ij}\cos(\omega_{ij} x + \phi_{ij})}_\text{Noise},
\end{equation}
where $A_{ij}$, $\omega_{ij}$ and $\phi_{ij}$ denote the noise amplitude, frequency and phase of the $j$th term of the $i$th example, and $\lambda_i$ denotes the average strength of the signal.
For simplicity, we take $\lambda_i$ to be constant and $x_i$ to be scalar here, but our results can easily be generalized.
With $\vec{A_i\omega_i} \equiv (A_{i1}, ..., A_{im})$, we can examine the gradient $\frac{\partial L_i}{\partial x}$, the key quantity in all gradient-based optimization.


\begin{boxedpreposition}
For a random $x_i$, the probability that $\frac{\partial L_i}{\partial x_i}$ points in the right direction approaches $\frac 1 2 + \frac 1 2 \text{erf}\left(\frac{\lambda_i}{|\vec{A_i\omega_i}|_2}\right)$ in the presence of multi-frequency noise.
\end{boxedpreposition}
\begin{proof}
(Full derivation in appendix~\ref{sec:app:theorems})
All constant terms, $\omega_{ij} = 0$, vanish in the gradient
$
\frac{\partial L_i}{\partial x_i} 
= \pm \lambda_i + \sum_{j=1}^m A_{ij} \omega_{ij} \sin(\omega_{ij} x_i + \phi_{ij}).
$
The probability density function (PDF) of the noise gradient is the convolution of the PDFs of the individual terms.
The PDF of a single term in the sum is $f_Y = \frac{1}{A\omega \pi \sqrt{1-(y/A\omega)^2}}$, which has zero mean and standard deviation $\sigma_{ij} = \frac{A_{ij}\omega_{ij}}{\sqrt 2}$.
For multi-frequency noise, the central limit theorem (CLT) is applicable and a good approximation.
Therefore, the PDF of the sum has $\sigma_i = \frac{|\vec{A_i\omega_i}|_2}{\sqrt 2}$.
The gradient is correctly aligned unless the noise component is stronger than the signal component $\lambda_i$.
Integrating the normal distribution yields $\frac 1 2 + \frac 1 2 \text{erf}\left(\frac{\lambda_i}{|\vec{A_i\omega_i}|_2}\right)$.
\end{proof}

We see that the gradient is likely to point towards $x_i^*$ when the signal-to-noise ratio $\text{SNR} = \frac{|\lambda_i|}{|\vec{A_i\omega_i}|_2}$ is high.
Note that the gradient noise scales not only with $A$ but also $\omega$, making high-frequency noise, i.e. chaos, especially problematic for optimization.

Now we can investigate the gradient quality for joint optimization.
We assume that all examples share a common $x^*$ and defer the general case to the next section. All other variables can vary between examples.
It turns out that, depending on the assumptions we make, there are two relevant ways of determining the update direction for $x$ based on the feedback from the individual examples.

\paragraph{Sum of losses:}
We optimize the total objective $L(x) = \sum_{i=1}^N L_i(x)$.

\begin{boxedtheorem2} \label{theorem:sum-of-losses}
    For equal noise levels across examples, the probability that the summed gradient direction is correct scales with $\sqrt N$.
\end{boxedtheorem2}
\vspace{-4mm}
\begin{proof}
(Full derivation in appendix~\ref{sec:app:theorems})
    The gradient is $\frac{\partial L}{\partial x} 
    = \pm \sum_i \lambda_i + \sum_{ij}  A_{ij} \omega_{ij} \sin(\omega_{ij} x + \phi_{ij})$.
    For the noise part, the CLT is applicable and yields a normal distribution with $\sigma = \frac{|\vec{A\omega}|_2}{\sqrt 2}$ for the PDF.
    Consequently, the probability that the gradient is correctly aligned is
    $\frac 1 2 + \frac 1 2 \text{erf}\left(\frac{\sum_i \lambda_i}{|\vec{A\omega}|_2}\right)$.
    If the noise level in all examples is equal, we can rewrite $|\vec{A\omega}|_2 = \sqrt N |\vec{A_i\omega_i}|_2$ for any $i$.
    Substituting the mean signal $\bar\lambda = \frac 1 N \sum_i \lambda_i$ yields
    $\frac 1 2 + \frac 1 2 \text{erf}\left(\frac{\bar\lambda \sqrt N}{|\vec{A_1\omega_1}|_2}\right)$ for the probability, which is equal to 
    $\frac 1 2 + \frac{\bar\lambda \sqrt N}{|\vec{A_1\omega_1}|_2 \sqrt \pi}$ up to second order.
\end{proof}

\paragraph{Majority vote:} Each example $i$ votes on whether $x$ should be increased or decreased.
\begin{boxedtheorem2} \label{theorem:vote}
    For equal SNRs across examples, the probability that the majority direction is correct scales with $\sqrt N$.
\end{boxedtheorem2}
\vspace{-4mm}
\begin{proof}
(Full derivation in appendix~\ref{sec:app:theorems})
    For equal gradient SNR, the probability that the majority direction is correct is given by the Bernoulli distribution,
    $\sum_{k > N/2} \binom{N}{k} (\frac 1 2 + \epsilon)^k (\frac 1 2 - \epsilon)^{n-k}$
    with $\epsilon = \frac{\text{SNR}}{\sqrt\pi}$.
    For large $N$, this term can be approximated as the integral
    $\int_{N/2}^\infty \mathcal N\left(\frac N 2 + N\epsilon, \sqrt{\frac N 4 - N \epsilon^2}\right) dx
    = \frac 1 2 + \frac 1 2 \text{erf}\left( \frac{\sqrt N \epsilon}{\sqrt{\frac 1 4 - \epsilon^2}} \right)$,
    which up to second order is equal to
    $\frac 1 2 + \frac{\sqrt N \epsilon}{\sqrt\pi \sqrt{\frac 1 4 - \epsilon^2}}$.
\end{proof}

Almost all machine learning techniques employ the first method of minimizing $\sum_i^N L_i$~\cite{DeepLearningBook}.
As demonstrated, this method improves approximately with $\sqrt N$ for similar noise levels across examples.
Under the altered assumption of similar SNRs, majority voting instead exhibits this $\sqrt N$ scaling while the summed objective loses this property.
However, there are several reasons for why $\sum_i^N L_i$ is almost always the method of choice.
For one, the assumption of equal noise is more prevalent, as most machine learning approaches are based on the assumption of independent and identically-distributed (i.i.d.) data~\cite{DeepLearningBook}.
Furthermore, the constant factors in the scaling laws derived above favor $\sum_i^N L_i$ in the sense that for equal SNRs \emph{and} equal absolute noise levels, minimizing the total objective is generally more efficient.
Lastly, in cases where the assumptions for $\sum_i^N L_i$ are strongly violated, researchers have employed other methods, such as gradient clipping~\cite{zhang2019gradient}, in order to keep learning stable.
Gradient clipping affects learning stability in a very similar way to majority voting and effectively constitutes a trade-off between the two methods.
For very low clipping thresholds, gradient clipping reduces the update vectors of all examples to the same magnitude, effectively assigning equal weight to every example, which exactly matches the behavior of majority voting.

\section{Joint parameterization of inverse problems with neural networks} \label{sec:reparameterization}

We have seen that the gradient accuracy in the presence of noise improves with $\sqrt N$ when jointly optimizing multiple inverse problems with the same optimum.
Now we consider the general case of different optima by adding a dependence on an example-specific vector, i.e. $x_i^* = x_i^*(\gamma_i)$.
Since this dependence is a priori unknown, a parameterized model capable of approximating the true dependence must be employed.
We refer to the optimization of this model as \emph{joint parameterized optimization} (JPO), and a common choice in previous work are neural networks $x_i = \mathcal N(\gamma_i, y_i \,|\, \theta)$ due to their ability to approximate arbitrary functions~\cite{DeepLearningBook}.
Inserting the model and summing the losses of all examples yields the jointly parameterized optimization problem
\begin{equation} \label{eq:reparameterized-problem-final}
    x_i^* = x_i(\theta^*)
    \quad\quad\quad
    \theta^* = \mathrm{arg min}_\theta \sum_{i=1}^n \mathcal L(F(\mathcal N(\gamma_i, y_i \,|\, \theta) \,|\, \gamma_i), y_i).
\end{equation}

\begin{wrapfigure}{r}{6cm}
    \centering
    \vspace{-2mm}
    \includegraphics[width=6cm]{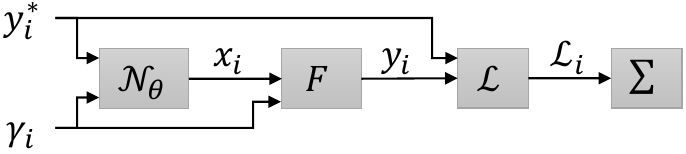}
    \caption{Parameterized optimization. The network predicts solutions $x_i$ based on targets $y_i^*$ and identifiers $\gamma_i$. The loss $\mathcal L$ is defined in the output space of $F(x)$.}
    \label{fig:sketch-reparameterized}
    \vspace{-5mm}
\end{wrapfigure}


Fig.~\ref{fig:sketch-reparameterized} shows the corresponding computational graph.
Solving this optimization problem equates to training the neural network $\mathcal N$ on the data set $\{ \gamma_i, y_i\}$ with the effective loss function $F \circ \mathcal L$.
%
Inserting the noise model yields 
\begin{equation}
    \frac{\partial L}{\partial\theta}
    = \sum_{i=1}^N \pm \lambda_i \frac{\partial x_i}{\partial\theta} + \sum_{i,j} A_{ij} \omega_{ij} \sin(\omega_{ij} x_i + \phi_{ij}) \frac{\partial x_i}{\partial\theta},
\end{equation}
where the factors $\frac{\partial x_i}{\partial\theta}$ map the gradient from the solution spaces to the shared weights $\theta$.
These factors enable cross-talk between the individual examples during the optimization, which can clearly be seen when considering steepest descent updates.
%
Then the steps up to first order in $x$ space for individual optimization and network training are
\begin{equation}\label{eq:JPO-update}
    \Delta x_i^\text{ind.} = - \eta \frac{\partial L_i}{\partial x_i}
    \quad\quad\quad\quad
    \Delta x_i^\text{JPO} = -\eta \sum_{j=1}^N \frac{\partial L_j}{\partial x_j} \frac{\partial x_j}{\partial \theta} \frac{\partial x_i}{\partial \theta} .
\end{equation}

With the network (JPO), the update direction depends on all examples, as indicated by the sum over $j$, enabling additional information to be used each step.
Without assuming a specific network architecture, we cannot exactly evaluate the factors $\frac{\partial x_i}{\partial\theta}$.
However, from a small set of assumptions, we can develop an empirical model for the network response.
We know that, for a large data set, any single network update is unlikely to lower the loss of all data points at the same time.
As an example, imagine a neural network fitting a high-frequency sine wave from $N$ sample points. 
Even without noise, the first update is unlikely to reduce the error of all $N$ samples, as the network prioritizes larger deviations first.
Nevertheless, the network will eventually fit all points perfectly if it is large enough.
This behavior is true even for simple tasks, such as fitting a linear function~\cite{DeepLearningBook}.

\begin{boxedtheorem2} \label{theorem:net-alignment}
    If aligning the update direction $\Delta x_N$ with the negative individual gradient $-\frac{\partial L_N}{\partial x_N}$ becomes exponentially less likely with $N$,
    the fraction of correct updates $\rho_N \rightarrow \xi \in [\frac 1 2, 1]$ as $N \rightarrow \infty$, with $\xi$ independent of the noise level.
\end{boxedtheorem2}
\vspace{-4mm}
\begin{proof}
(Full derivation in appendix~\ref{sec:app:net-model})
Formalizing the assumption, we can express the probability of the $N$th example receiving an update aligned with its negative gradient as $\exp\left(- \frac{N-1}{A}\right)$, where $A$ is an inherent property of the network architecture referred to as its plasticity~\cite{lyle2023plasticity}.
We also need to take into account that the update can be aligned even if the gradient is not factored into the network update.
This probability depends on how close it is to another example and on the complexity $C$ of the function being fitted.
For random solution predictions $x_i$, the probability of a new example being independent is $C \frac{D}{N}$ up to first order, where $D$ denotes the size of the support of the distribution of $x_i$.
With $\tilde C \equiv CD$, the probability of an aligned update for an isolated example being $\frac 1 2$, and knowing that identical examples receive the same update, we can write the update alignment probability as
\vspace{-2mm}
\begin{equation} \label{eq:plasticity-complexity-model}
P(N\text{th aligned}) = 
    \left( 1 - e^{- \frac{N-1}{A}} \right)
    \left( \frac{C}{2N} + \left(1 - \frac C N\right) \rho_{N-1} \right)
    + e^{- \frac{N-1}{A}},
\end{equation}
\vspace{-6mm}

where $\rho_{N-1}$ denotes the previous fraction of aligned updates.
This gives us the recursive formulation $\rho_N = \frac 1 N ( (N-1)\rho_{N-1} + P(N\text{th aligned}) )$, $\rho_1 = 1$ which allows us to predict the expected fraction of correct updates based on the noise level, problem complexity $C$, and network plasticity $A$.
Integrating Eq.~\ref{eq:plasticity-complexity-model} yields that $\rho_N \xrightarrow{N\rightarrow \infty} \xi(\tilde C, A) \in [\frac 1 2, 1]$.
\end{proof}

\textit{Note:} The assumed exponential decrease in fitting capability $e^{-(N-1)/A}$ can be interchanged for other functions as long as they approach zero as $N\rightarrow\infty$. We used this form because it is simple and matches experimental data quite well.

Theorem~\ref{theorem:net-alignment} tells us that the fraction of examples that get an aligned gradient decreases from 1 as we increase $N$, approaching a constant $\xi$ between $0.5$ and 1.
This counteracts the $\sqrt{N}$ increase we get from theorems \ref{theorem:sum-of-losses} and \ref{theorem:vote}, resulting in a more involved overall dependence which still approaches $\xi$ in the limit $N\rightarrow\infty$.
This reflects the fact that an increasing number of optimization steps are required to fit larger data sets with first-order methods like SGD or Adam~\cite{Adam}, as individual optimization steps cannot optimize all examples at the same time.
We show example curves in section~\ref{sec:curve-fitting}.
The value of $\xi$ represents the case where all examples will converge if the network is capable enough and trained for long enough.
%
The $\sqrt{N}$ dependence therefore only applies to the solutions after the network has been fully trained.

However, from Eq.~\ref{eq:JPO-update} we see that for $N\rightarrow\infty$, the noisy case approaches the noise-free case for which we know that a sufficiently large network can approximate the desired solutions to arbitrary precision~\cite{NNApproximator}.
Thus the network can find the optimal minima in the case of infinite training data, avoiding local minima and progressing through chaotic regions.
With end-to-end training, this constitutes a method of finding optimal solutions on the network's training set, with the network training taking the place of the inverse problem optimizer.
Consequently neural networks can also be used as replacements or additional components in solving a set of known inverse problems, in which case generalization to new data is of secondary importance.
This benefit of joint optimization has not been explored so far, as networks have mostly been used for fast inference on unseen data.
When the goal is solving known inverse problems, there is no need to use early stopping, because overfitting to the training data is desired.
We validate these results experimentally in section~\ref{sec:experiments}.



\myvspace{}
\paragraph{Limitations}
In this work, we focus on unstructured solution spaces. As mentioned above, related work has studied structured spaces, such as images or grid-based simulations, and specific network architectures can be used to exploit the structure.
Furthermore, we only consider unconstrained optimization problems, enforcing hard constraints by running bounded parameters through a scaled $\tanh$ function which naturally clamps out-of-bounds values in a differentiable manner.
While neural network inference is typically faster than running a classical optimizer, an additional training stage is required for the network.
For a high desired accuracy and small data set size $N$, optimizing a neural network with first-order optimizers takes longer than optimizing the examples individually with a high-performance classical optimizer.
While this trend reverses for large $N$, we specifically choose small $N$ to investigate the scaling behavior.
Wall-clock training times are provided in the appendix.



\section{Experiments} \label{sec:experiments}

The previous sections showed that neural networks are able to find better solutions to inverse problems as the data set size $N$ increases.
We now test the theory by performing a series of challenging numerical experiments, both under controlled conditions as well as on real-world PDEs.
To save computational resources, we switch to individual higher-order optimization once the network approaches a fixed minimum on all examples.
We use gradient clipping at the 90th percentile for all experiments
and train off-the-shelf neural network architectures with Adam~\cite{Adam}, adjusting only the learning rate to achieve stable convergence.
This intentional lack of hyperparameter tuning ensures that our results reflect the capabilities of JPO as a black-box extension of generic optimizers.
We run each experiment and method multiple times, varying network initialization and data sets.
Additional details can be found in Appendix~\ref{app:experiments}, along with supervised training and the neural adjoint method as additional baselines.
An additional experiment replicating a setup from previous work~\cite{NeuralAdjoint} is given in Appendix~\ref{app:additional-experiments}.

\begin{wrapfigure}{r}{7cm}
    \centering
    \vspace{-4mm}
    \includegraphics[width=\linewidth]{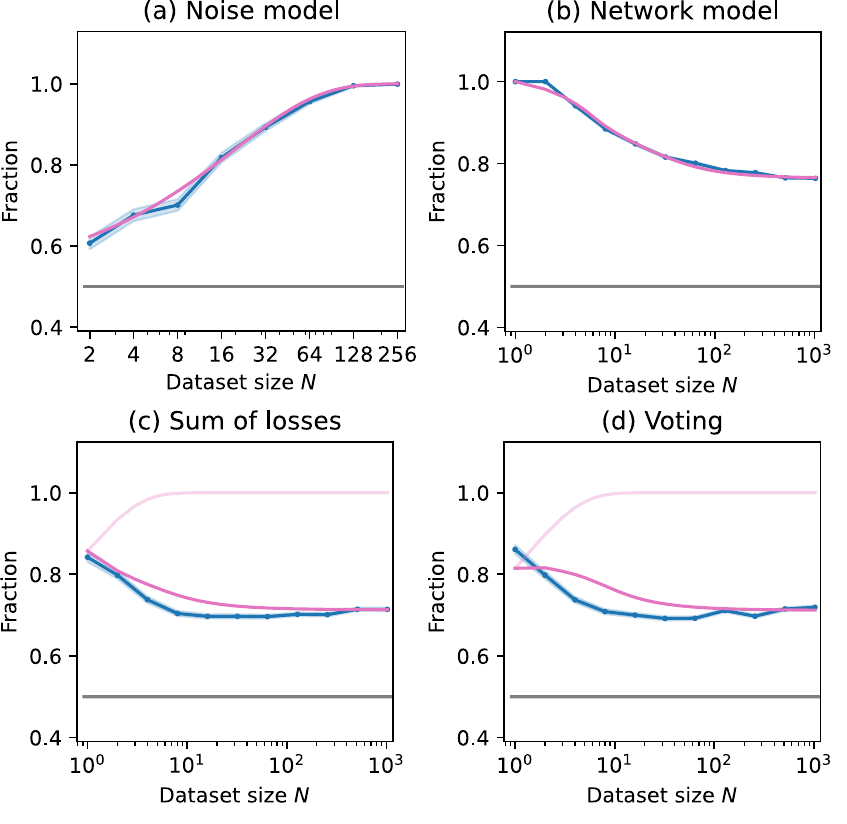}
    \vspace{-6mm}
    \caption{Initial gradient alignment with JPO. Measured average over 1000 seeds (blue) and theory curve (pink).
    \textbf{(a)}~Noise scaling when examples have the same ground truth $x^*$.
    \textbf{(b)}~Noise-free network training with $x^*_i = 10 \gamma_i$. 
    \textbf{(c,d)}~Sum-of-losses and voting methods on the problem of fitting $\sin(2x)$ in the presence of noise.
    Noise-only theory curve in light pink.
    }
    \label{fig:models}
    \vspace{-6mm}
\end{wrapfigure}


\paragraph{Synthetic problems with controlled noise.} \label{sec:curve-fitting}
To test the predictions from theorems \ref{theorem:sum-of-losses}, \ref{theorem:vote} and \ref{theorem:net-alignment}, we perform a series of experiments with artificial noise as in Eq.~\ref{eq:fourier-model}.
First, we can compare the theory predictions derived in section~\ref{sec:joint-optimization} to measurements from controlled experiments.
We construct 1000 cases for various $N$ and compute the gradient at random points in the loss landscape.
The predicted and measured scaling behavior for $\sum_i L_i$ (Fig.~\ref{fig:models}a) match perfectly, and majority voting exhibits identical behavior. 

Next we test the network response model in isolation, letting the solutions $x_i^* = 10 \gamma_i$ depend linearly on an example parameter $\gamma_i$.
We use SGD to update $\theta$ and then run the network again to obtain the update direction for each example.
Fig.~\ref{fig:models}b shows that the predictions for $\tilde C=6.4, A=12.9$ closely match the actual behavior of the network.

Finally, we test JPO on the noisy problem of learning $\sin(2x)$, $x \in [-2, 2]$ (Fig.~\ref{fig:models}c).
Since the plasticity $A$ and complexity $\tilde C$ do not depend on the noise, we fit them on the noise-free case beforehand.
The model reproduces the behavior of the data well, especially for large $N$.
We see that increasing $N$ generally leads to less correct gradients, despite the positive statistical influence, indicating that more training iterations are required for larger $N$.
Tracking gradients alone is therefore insufficient for predicting a model's final performance, and we will consider fully trained networks in the following experiments.
%



\paragraph{Wave packet localization}

\begin{figure*}[tb]
    \centering
    \includegraphics[width=\textwidth]{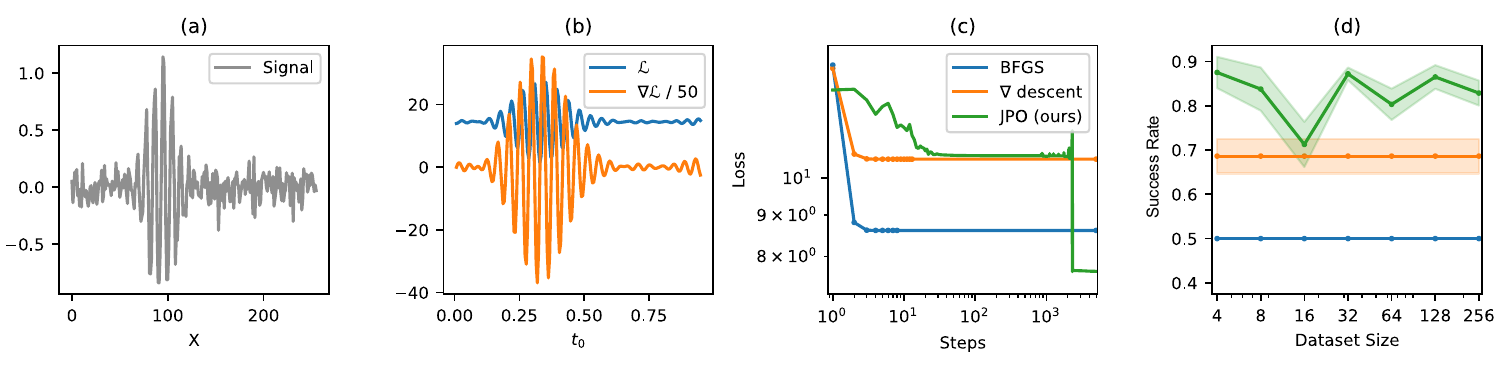}
    \vspace{-8mm}
    \caption{Wave packet localization.
    \textbf{(a)}~Example waveform $u(t)$,
    \textbf{(b)}~Loss and gradient landscape for $t_0$, \textbf{(c)}~training / optimization curves, \textbf{(d)}~Fraction of inverse problems for which JPO and gradient descent yield better solutions than BFGS.}
    \label{fig:wavepacket}
\end{figure*}

Next we look at an inverse problem with local minima.
The task consists of locating the center $t_0$ of a wave packet from a noisy signal $u(t)$ that is sampled at $t = 1, ..., 256$.
An example waveform is shown in Fig.~\ref{fig:wavepacket}a, and
Fig.~\ref{fig:wavepacket}b shows the local minima in the loss landscape.
%
A perfect solution with $\mathcal L = 0$ is impossible due to the noise.
The optimization curves (Fig.~\ref{fig:wavepacket}c) reveal that BFGS and gradient descent often converge to local minima and terminate.
BFGS usually takes a large step in the first iteration and then quickly converges to the closest optimum while gradient descent takes more iterations to converge.
Which method finds the better solution is somewhat random, but in this experiment gradient descent comes closer to the global optimum more often, likely because it is less prone to performing overly large update steps.

To parameterize the problem, we create a neural network $\mathcal N$ that maps the 256 values of the observed signal $u(t)$ to the unknown value $t_0$.
We use a standard architecture~\cite{VGG2014} and train it according to Eq.~\ref{eq:reparameterized-problem-final}.
%
During the optimization, the estimate of $t_0$ repeatedly moves from minimum to minimum until settling after 500 to 3000 iterations.
Many examples do not converge to the global optimum, but the cross-talk between different examples, induced by the shared parameters $\theta$, regularizes the movement in $t_0$ space, preventing solutions from moving far away from the global optimum.
%
Fig.~\ref{fig:wavepacket}d shows the results for different numbers of inverse problems $N$.
Since BFGS optimizes each example independently, the data set size has no influence on its performance.
Overall, JPO finds better solutions than BFGS in around 80\% of examples.
This problem can also be successfully solved with supervised learning, see appendix~\ref{app:wavepacket}.

\myvspace{} 
\paragraph{Billiards}

\begin{wrapfigure}{r}{7cm}
    \centering
    \vspace{-6mm}
    \includegraphics[width=\linewidth]{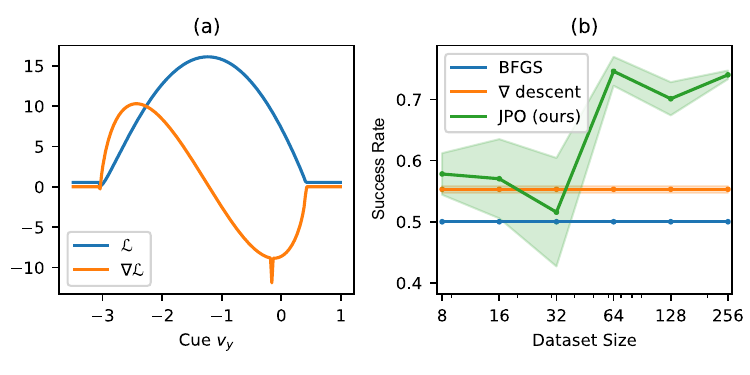}
    \vspace{-6mm}
    \caption{Billiards experiment.
    \textbf{(a)}~Example loss landscape,
    \textbf{(b)}~Fraction of inverse problems for which JPO and gradient descent yield better solutions than BFGS.}
    \label{fig:billiards}
\end{wrapfigure}

Next, we consider a rigid-body setup inspired by differentiable billiards simulations of previous work \cite{hu2019difftaichi}.
The task consists of finding the optimal initial velocity $\vec v_0$ of the cue ball to cause a non-elastic collision and propel another ball towards a target location (Fig.~\ref{fig:billiards}a).
A collision only occurs if $\vec v_0$ is large enough and pointed towards the other ball.
Otherwise, the second ball stays motionless, resulting in a constant $\mathcal L$ and $\frac{\partial \mathcal L}{\partial \vec v_0} = 0$.
This causes classical optimizers to fail unless initialized close to the correct solution.

Employing a fully-connected neural network with JPO drastically improves the solutions.
While for $N \leq 32$ only small differences to BFGS can be observed, access to more inverse problems lets gradients from some problems steer the optimization of others that get no useful feedback.
This results in almost all problems converging to the solution for $N \geq 64$ (see Fig.~\ref{fig:billiards}b).

\myvspace{} 
\paragraph{Kuramoto–Sivashinsky equation}

\begin{figure*}[tbh]
    \centering
    \includegraphics[width=\textwidth]{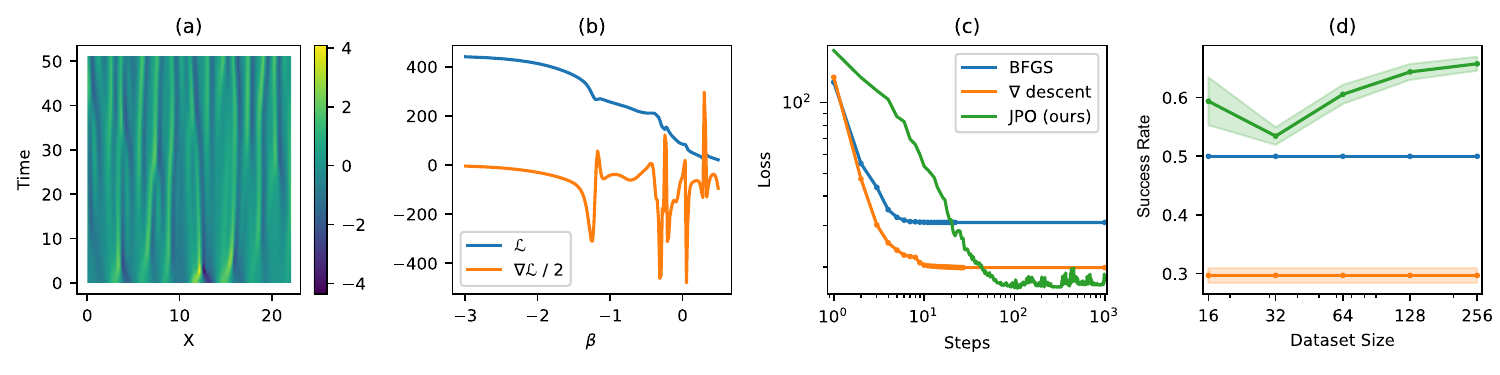}
    \vspace{-8mm}
    \caption{Kuramoto–Sivashinsky experiment. \textbf{(a)}~Example trajectory,
    \textbf{(b)}~loss landscape for $\beta$,
    \textbf{(c)}~example optimization curves,
    \textbf{(d)}~fraction of inverse problems for which JPO and gradient descent yield better solutions than BFGS.}
    \label{fig:ks}
\end{figure*}

Originally developed to model the unstable behavior of flame fronts~\cite{KS1978}, the Kuramoto–Sivashinsky (KS) equation models a chaotic system, $\dot u(t) = -\frac{\partial^2 u}{\partial x^2} - \frac{\partial^4 u}{\partial x^4} - u\cdot \nabla u$.
We consider a two-parameter inverse problem involving the forced KS equation
\begin{equation*}
    \dot u(t) = \alpha \cdot G(x) -\frac{\partial^2 u}{\partial x^2} - \frac{\partial^4 u}{\partial x^4} - \beta \cdot u\cdot \nabla u,
\end{equation*}
where $G(x)$ is a fixed time-independent forcing term and $\alpha,\beta \in \mathbb R$ denote the unknown parameters governing the evolution.
Each inverse problem starts from a randomly generated initial state $u(t=0)$ and is simulated until $t=25$, by which point the system becomes chaotic but is still smooth enough to allow for gradient-based optimization.
Fig.~\ref{fig:ks}a shows an example trajectory, and the corresponding loss landscape $\frac{\partial \mathcal L}{\partial \beta}\|_{\alpha=\alpha^*}$ at the true value of $\alpha$ is shown in Fig.~\ref{fig:ks}b.
Despite the chaotic nature of the loss landscape, BFGS manages to find the correct solution in about 60\% of cases, beating gradient descent which gets stuck in local optima more frequently.
Fig.~\ref{fig:ks}c shows the optimization curves for finding $\alpha,\beta$.

For parameterization, we use a neural network consisting of 2D convolutional layers followed by fully-connected layers, similar to the wave packet experiment.
Solving the inverse problems with JPO sees over 80\% of examples approach the global minimum.
However, without refinement the accuracy stagnates far from machine precision.
Refining these solutions with BFGS decreases the remaining error of these cases to machine precision in 4 to 17 iterations, less than the 12 to 22 that BFGS requires when initialized from the mean ground truth solution $\mathbb E[P(\xi)]$.
Again, we observe that the performance increases with $N$ as expected.
For $N=256$, JPO finds better solutions than BFGS in 65\% of cases.

\myvspace{} 
\paragraph{Incompressible Navier-Stokes}


\begin{figure*}[htb]
    \centering
    \includegraphics[width=\textwidth]{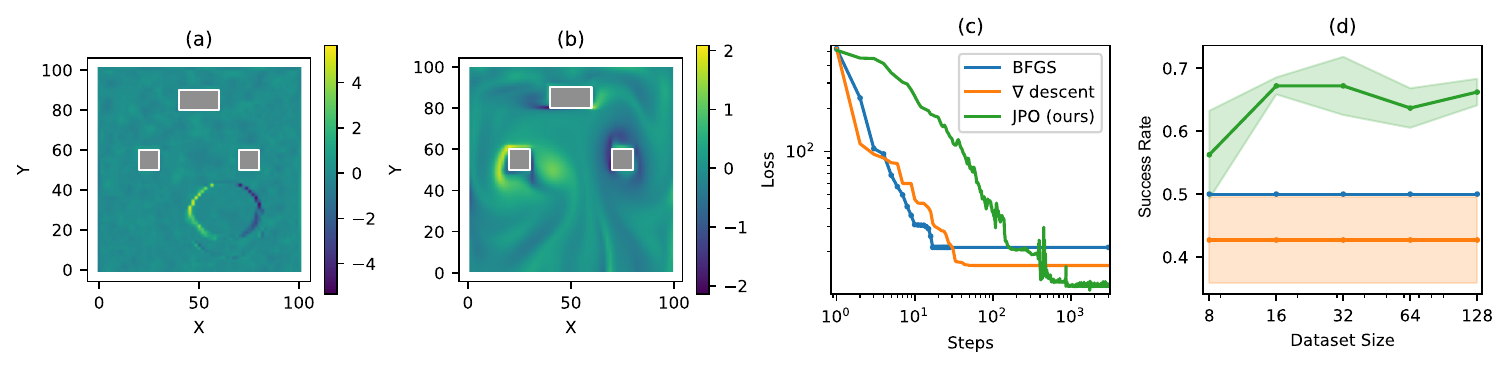}
    \vspace{-8mm}
    \caption{Fluid experiment.
    \textbf{(a,b)}~example initial and final velocity field, obstacles in gray. Only $y \geq 50$ is observed.
    \textbf{(c)}~example median optimization curves,
    \textbf{(d)}~fraction for which JPO and gradient descent yield better solutions than BFGS.}
    \label{fig:fluid}
\end{figure*}

Incompressible Newtonian fluids are described by the Navier-Stokes equations,
\begin{align*}
    \dot u(\vec x, t) = \nu \nabla^2 u - u \cdot \nabla u - \nabla p
    \quad
    \mathrm{s.t.}
    \quad
    \nabla \cdot v = 0
\end{align*}
with $\nu \geq 0$, and their solutions can be highly complex~\cite{FluidsBook1967}.
Incompressible fluids represent a particularly challenging test case but one that is relevant for a variety of real-world problems~\cite{pope2000_Turbulent}.
We consider a setup similar to particle image velocimetry~\cite{PIV1997} in which the velocity in the upper half of a two-dimensional domain with obstacles can be observed.
The velocity is randomly initialized in the whole domain and a localized force is applied near the bottom of the domain at $t=0$.
The task is to reconstruct the position $x_0$ and initial velocity $\vec v_0$ of this force.
However, only the initial and final velocity in the top half of the domain is observable.
The initial velocity in the bottom half is unknown.
Fig.~\ref{fig:fluid}a,b show an example initial and final state of the system.
The final velocity field is measured at $t=56$ by which time small-scale eddies have merged into larger ones and the obstacles have had a major impact on the flow.

Fig.~\ref{fig:fluid}c shows the optimization curves.
On this problem, BFGS converges to some optimum in all cases, usually within 10 iterations, sometimes requiring up to 40 iterations.
Still, many examples get stuck in local optima.

For joint optimization, we parameterize the solution space using a network architecture featuring four 2D convolutional layers and two fully-connected layers.
For all tested $N$, JPO produces larger mean loss values than BFGS, especially for small $N$.
This results from about 10\% of examples seeing higher than average loss values. 
Nonetheless, 66.7\% of the inverse problems are solved more accurately than BFGS on average for $N>4$.

\section{Discussion}

Overall we find that using off-the-shelf neural network architectures and optimizers with no tuning to the specific problem, JPO finds better solutions than BFGS, the best tested iterative solver, in an average of 69\% of tested problems. 
Our theoretical computations as well as experimental results clearly show that this fraction increases with the data set size $N$.
Meanwhile, the computational cost of training the network increases sub-linearly with $N$, i.e. slower than with classical optimization.
This makes optimizing large numbers of inverse problems with JPO desirable, as better solutions can be obtained at less computational cost than with iterative optimizers, while also producing a predictor for unseen data.

Especially for small $N$, JPO sometimes yields worse solutions than iterative optimization.
This is unavoidable on the non-convex problems we study, since the limited statistics from small $N$ allow gradients to veer off towards local minima, as described in section~\ref{sec:joint-optimization}.
However, many of these cases can easily be identified by their outlier loss values, and falling back to iterative optimizers would further alleviate this issue.
On convex optimization problems, which have been extensively studied by previous work~\cite{NeuralAdjoint}, JPO always converges to the global optimum, as we show in an additional experiment in appendix~\ref{app:additional-experiments}.

We have additionally performed our experiments with two related machine-learning techniques, detailed in appendix~\ref{app:experiments}.
(i) Supervised training which can be used for unimodal problems with known ground truth.
Due to the multi-modal nature of the problems tested here, supervised learning yields worse results than JPO in all tested experiments (see Tab.~\ref{tab:app:better-fractions}), only beating BFGS in more than half of the examples on the wave packet experiment.
When coupled with a secondary BFGS optimization, it is competitive in our fluid experiment (Tab~\ref{tab:better-fractions}) but loses the sub-linear performance scaling.
(ii) The neural adjoint method~\cite{NeuralAdjoint} offers a way of finding solutions without requiring differentiable forward problems.
However, this method also scales poorly to our challenging problems.

\section{Conclusions and Outlook}

We have investigated the effects of joint optimization of multiple inverse problems when using neural networks to predict solutions directly.
We have shown that, in ideal circumstances, noise in the objective is suppressed with the square root of the number of jointly optimized problems, which aids convergence in the presence of local minima, chaos, or zero-gradient regions.
We have motivated the use of gradient clipping as a soft way of including majority voting, which we have shown to be the appropriate choice of gradient reduction for similar SNRs.
Our experiments show that the theoretical scaling translates to complex physics-based scenarios, where individual loss landscapes can be extremely challenging.
In all tested settings, JPO surpassed the individual optimizations in terms of solution quality when given enough examples, and its relative performance also increases for large data sets.

Tuning the network architectures to the specific problems could lead to further improvements in performance but would make the approach domain-dependent.
We have explicitly not performed architectural tuning beyond simple learning rate optimization to ensure that our results hold for a large variety of problems.
With no more information required than classical optimizers, JPO can be used as a generic drop-in component for existing optimization libraries.
An implementation would internally set up a standard neural network with the required number of inputs and outputs, then run the optimization with the largest stable learning rate.
This approach hides details of the training process, network architecture, and hyperparameters from users while making the gains in optimization accuracy conveniently accessible.
Our source code implements these concepts and will be made public in order to facilitate adoption.

A possible extension of our work is translating our results to probabilistic approaches, where instead a set of optima is predicted~\cite{mishra2020bayesian}.
From accelerating matrix multiplications~\cite{LearnedMatmul2022} to solving systems of linear equations~\cite{NNPreconditionDirac2023,NNPreconditionCG2019}, it is becoming increasingly clear that machine learning methods can be applied to purely numerical problems outside of typical big data settings,
and our results show that this also extends to solving nonlinear inverse problems.





\clearpage



\bibliographystyle{unsrtnat}
\bibliography{bibliography}


\clearpage
\appendix

\section{Derivation of the Theorems} \label{sec:app:theorems}

In this section, we derive and explain the theorems and proofs from section~\ref{sec:joint-optimization} of the main paper in more detail.
We first derive the properties of the Fourier noise model, starting with a single sine wave before moving on to sums of sine waves with different amplitudes, frequencies and phases.
We then derive the theorems for majority voting and summing the individual loss functions.

\subsection{Noise Models}
Here we examine the case that the gradient of optimization problems contains one part that points toward a good solution, the \emph{signal}, and the rest, which we refer to as \emph{noise}.
The noise is what prevents the optimizer from converging directly to the solution, e.g. by introducing additional local minima to the optimization landscape.

\paragraph{Sine noise}

Let's start with the noise being modeled by a sine wave parameterized by frequency $\omega$, amplitude $A$ and phase $\phi$.
Consider individual optimization problems defined as the minimization of $L_i(x)$
$$L_i(x) = \underbrace{\lambda_i |x_i - x_i^*|}_\text{Signal} \underbrace{-A_i\cos(\omega_i x + \phi_i)}_\text{Noise}$$
where $x_i^*$ is the ground-truth solution but not necessarily the exact minimum point.
The specific form of the good loss 
The gradient w.r.t. $x$ is
$$\frac{\partial L_i}{\partial x_i} = \pm \lambda_i + A_i \omega_i \sin(\omega_i x + \phi_i)$$
For random $\phi_i$, this gradient is aligned with the ground truth gradient $\pm \lambda_i$ with probability
$$P\left(\frac{\partial L_i}{\partial x_i} \sim (x_i-x_i^*)\right) 
= \frac{1}{2} + \frac{\sin^{-1}(\lambda_i / A_i \omega_i)}{\pi}
\approx \frac 1 2 + \frac{\lambda_i}{\pi A_i \omega_i}
$$
with the series expansion $\sin^{-1}(x) = x + \mathcal O(x^3)$.
This can be shown easily by integrating one sine period.
For the case of only noise, $\lambda_i=0$, the probability of the correct gradient is 50\%, and it increases up to 1 as the signal-to-noise ratio improves.

\paragraph{Fourier noise model}
The sine noise model is not very realistic for practical applications.
Instead, we model the noise distribution via its Fourier series, as this decomposition imposes few requirements on the noise.
Any piecewise continuous function with a finite number of extrema and discontinuities can be approximated arbitrarily accurately using the Fourier series.
This results in the optimization objective
$$L_i(x) = \underbrace{\lambda_i |x_i - x_i^*|}_\text{Signal} \underbrace{\sum_j -A_{ij}\cos(\omega_{ij} x + \phi_{ij})}_\text{Noise}$$
where the index $j$ refers to the $j$th term in the Fourier series.
To derive the probability of the gradient pointing in the right direction, we need to estimate the probability density function (PDF) of the noise.
The PDF of a single sine $A\omega\sin(\omega x + \phi)$ is
$$f_Y = \frac{1}{A\omega \pi \sqrt{1-(y/A\omega)^2}}$$
which has mean $\mu=0$ and standard deviation 
$$\sigma = \sqrt{\text{Var}[Y]} = \sqrt{\int_{-1}^1 \frac{y^2}{A\omega \pi \sqrt{1-(y/A\omega)^2}} dy} = \frac{A\omega}{\sqrt 2}.$$
The PDF of the sum of multiple random variables is given by the convolution of their PDFs if they are independent.
This can technically be done for the sines but is extremely ugly and not insightful (WolframAlpha can compute the indefinite convolution but not the definite one in time).

However, for many summation terms, we can use the central limit theorem to approximate the combined PDF as a normal distribution $\frac{1}{\sigma\sqrt{2\pi}} e^{-x^2/2\sigma^2}$ with standard deviation
$$\sigma_i = \sqrt{\sum_j \sigma_{ij}^2} = \sqrt{\sum_j \frac{A_{ij}^2\omega_{ij}^2}{2}} = \frac{|\vec{A_i\omega_i}|_2}{\sqrt 2},$$
where $\vec {A\omega}$ is the vector of the Fourier components multiplied by their respective frequencies.
Based on the normal distribution, the probability that the gradient points in the right direction is
$$P\left(\frac{\partial L_i}{\partial x_i} \sim (x_i-x_i^*)\right) 
= \frac 1 2 + \frac 1 2 \text{erf}\left(\frac{\lambda_i}{\sqrt 2 \sigma_i}\right) 
= \frac 1 2 + \frac 1 2 \text{erf}\left(\frac{\lambda_i}{|\vec{A_i\omega_i}|_2}\right) 
\approx \frac 1 2 + \frac{\lambda_i}{\sqrt{\pi} |\vec{A_i\omega_i}|_2}$$
with the series expansion $\text{erf}(x) = \frac{2x}{\sqrt\pi} + \mathcal O(x^3)$.
This is similar to the result we obtained for the sine noise model above, but the error function leads to slightly different scaling factors than the $\sin^{-1}$ to account for the more complex model which can result in narrower spikes in the loss landscape.

\includegraphics[width=6cm]{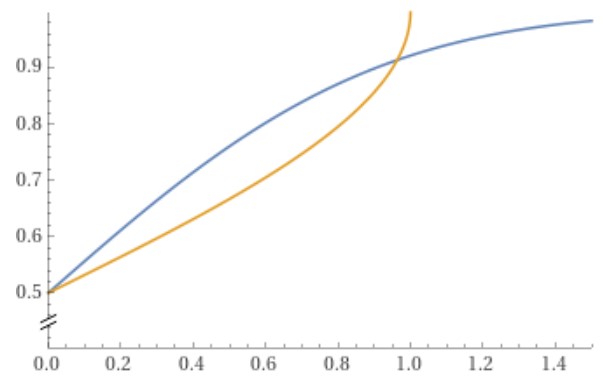}

\subsection{Joint Optimization}
We now optimize multiple independent examples $i$ jointly, replacing $x_i \equiv \hat x_i(\theta)$ by a model that can capture the true dependence.
Let's start with the simple model $\hat x_i(\theta) = \theta$, i.e. all $x_i$ are equal and all examples share the same minimum $x^*_i = \theta^*$.
We can consider two ways of obtaining an update direction for $\theta$.

\paragraph{Majority vote}
We can perform a majority vote for each parameter in $\theta$ to determine whether it should be increased or decreased.
Assuming equal gradient signal-to-noise ratios $\epsilon = \frac{|\lambda_i|}{\sqrt\pi |\vec{A_i\omega_i}|_2} > 0$, this yields the probability that the majority direction is correct of
$$P\left(\frac{\partial L}{\partial\theta} \sim (\theta - \theta^*)\right)
= \sum_{k > N/2} \binom{N}{k} (\frac 1 2 + \epsilon)^k (\frac 1 2 - \epsilon)^{n-k}.
$$
For large $N$, this term can be approximated as a normal distribution, yielding
$$P\left(\frac{\partial L}{\partial\theta} \sim (\theta - \theta^*)\right)
= \int_{N/2}^\infty \mathcal N\left(\frac N 2 + N\epsilon, \sqrt{\frac N 4 - N \epsilon^2}\right) dx
= \frac 1 2 + \frac 1 2 \text{erf}\left( \frac{\sqrt N \epsilon}{\sqrt{\frac 1 4 - \epsilon^2}} \right)
\approx \frac 1 2 + \frac{\sqrt N \epsilon}{\sqrt\pi \sqrt{\frac 1 4 - \epsilon^2}}
$$
which increases with the square of the number of optimized examples $N$.

\paragraph{Sum of Losses}
Another possibility is defining a new objective $L$ to be optimized for:
$$L(\theta) = \sum_i L_i(\hat x_i(\theta)).$$
Differentiating $L(\theta)$ w.r.t. the model parameters $\theta$ yields
$$
\frac{\partial L}{\partial\theta} 
= \sum_i \left( \pm \lambda_i + \sum_j A_{ij} \omega_{ij} \sin(\omega_{ij} x + \phi_{ij}) \right)
= \pm \sum_i \lambda_i + \sum_{ij}  A_{ij} \omega_{ij} \sin(\omega_{ij} x + \phi_{ij}),
$$
where we have used that all $\pm \lambda_i$ have the same sign.
The gradient is aligned with the true gradient if the second sum does not overpower the $\lambda$-sum.
We estimate the PDF of the second sum using the central limit theorem, recalling that all terms have mean $\mu_i=0$ and standard deviation $\sigma_i =\frac{|\vec{A_i\omega_i}|_2}{\sqrt 2}$.
Then, the PDF of the sum is a normal distribution with $\mu=0$ and
$$\sigma = \sqrt{\sum_i \sigma_i^2} 
=\sqrt{\sum_i \frac{|\vec{A_i\omega_i}|_2^2}{2}} 
= \frac{|\vec{A\omega}|_2}{\sqrt 2}$$
where $|\vec{A\omega}|_2 = \sqrt{\sum_{ij} A_{ij}\omega_{ij}}$ since the norm of norms is the norm of all values.
Consequently, the probability that the gradient is correctly aligned is
$$
P\left(\frac{\partial L}{\partial\theta} \sim (\theta - \theta^*)\right)
= \frac 1 2 + \frac 1 2 \text{erf}\left(\frac{\sum_i \lambda_i}{\sqrt 2 \sigma}\right) 
= \frac 1 2 + \frac 1 2 \text{erf}\left(\frac{\sum_i \lambda_i}{|\vec{A\omega}|_2}\right)
$$
If the noise in all examples is roughly equally distributed, we can approximate $|\vec{A\omega}|_2 \approx \sqrt N |\vec{A_1\omega_1}|_2$ (or any other $i$).
Also inserting the mean true slope $\bar\lambda = \frac 1 N \sum_i \lambda_i$ yields
$$
P\left(\frac{\partial L}{\partial\theta} \sim (\theta - \theta^*)\right)
\approx \frac 1 2 + \frac 1 2 \text{erf}\left(\frac{\bar\lambda \sqrt N}{|\vec{A_1\omega_1}|_2}\right)
\approx \frac 1 2 + \frac{\bar\lambda \sqrt N}{|\vec{A_1\omega_1}|_2 \sqrt \pi}
$$

Evidently majority voting and sum of losses show the same accuracy scaling with $\sqrt N$.
However, the assumptions that lead to these results are different.
For the majority voting, we assumed equal signal-to-noise ratio while for the sum of losses we assumed equal noise levels.
This fact sheds some light onto which method should perform better given some noise distribution.
If the noise is approximately equal for all examples, the sum of losses should be preferred, but if the noise level can vary drastically, methods similar to majority voting should be more robust.

In practice, we deal with this by performing gradient clipping on the sum-of-losses method.
By adjusting the percentile cutoff, we can effectively interpolate between sum-of-losses and majority voting.
The lower the clipping threshold, the closer the method becomes to majority voting.

\subsection{Multivariate models}

Next we look at the more general case that the examples do not share the same minimum point, starting with a linear dependence.
As before, we assume that the functional form of this dependence is known and, thus, the task consists of finding $\theta^*$ by optimizing $\theta$.
With $\theta$-dependence, the combined signal and noise loss is
$$
L_i(x) = \underbrace{\lambda_i |x_i(\theta) -  x_i^*| }_\text{Signal} + \underbrace{\sum_j-A_{ij}\cos(\omega_{ij} x_i(\theta) + \phi_{ij})}_\text{Noise},
$$
and the corresponding gradient is
$$
\frac{\partial L_i}{\partial\theta} 
= \pm \lambda_i \frac{\partial x_i}{\partial\theta} + \sum_j A_{ij} \omega_{ij} \sin(\omega_{ij} x_i + \phi_{ij}) \frac{\partial x_i}{\partial\theta}
$$
where both signal and noise term now scale with $ \frac{\partial x_i}{\partial\theta}$.


\subsection{Generalizations}

Here we assumed the true gradient direction to be contained as $\lambda_i |x_i^* - x_i|$.
This is of course not the case in most settings.
Instead, magnitude of the good gradient may vary across $x$ and may not point directly toward the minimum.
Considering the varying magnitude, this can easily be added to our model by making $\lambda_i$ a random variable.
When including directions that do not point towards a minimum, we have to make assumptions about the optimization in the absence of noise and then show that the noisy optimization approximately yields the same updates.
Here, the argument becomes more difficult to interpret as a reduction in loss is also achieved by the non-parameterized optimizations, only towards potentially non-optimal points.

\clearpage
\section{Network Update Model} \label{sec:app:net-model}

In section~\ref{sec:reparameterization} of the paper, we introduce the model we use to estimate the fraction of examples that receive a correct update in the absence of noise.
For one example $i$, the steepest descent direction is $-\frac{\partial L_i}{\partial x_i}$.
However, the update direction with JPO (Eq.~\ref{eq:JPO-update}) is
$\Delta x_i^\text{JPO} = -\eta \sum_{j=1}^N \frac{\partial L_j}{\partial x_j} \frac{\partial x_j}{\partial \theta} \frac{\partial x_i}{\partial \theta}$ which may or may not be aligned with steepest descent.
For a 1D solution space per example, the update aligns with the steepest descent directions if both have the same sign.

To arrive at the final functional form of the model, we consider adding the $N$th example to the data set and start from the updates resulting from optimizing the other $N-1$ examples only.
Since we know that for $N=1$ the update is guaranteed to be aligned with the negative descent, we can derive the probabilities for all examples by induction ($N \rightarrow N+1$).

We split the probability that $\Delta x_N^\text{JPO}$ is aligned with $-\frac{\partial L_N}{\partial x_N}$ into multiple branches:
\begin{enumerate}
    \item If the update $\Delta x_N^{\text{JPO w/o }N} = -\eta \sum_{j=1}^{N-1} \frac{\partial L_j}{\partial x_j} \frac{\partial x_j}{\partial \theta} \frac{\partial x_N}{\partial \theta}$ is already aligned with $-\frac{\partial L_N}{\partial x_N}$, the probability of alignment is 1.
    \item Else the model needs to adjust the update. The probability of alignment is $e^{-\frac{N-1}{A}}$.
    \item Case 1 occurs with 50\% probability if the example is uncorrelated to all previous examples and with probability $\rho_{N-1}$ if it is fully correlated with another example.
    \item The probability that example $N$ is correlated scales with its distance to the other examples and the problem complexity $C$. The distance to other examples scales inversely proportional to $N$, independent of the dimensionality of the solution space. Therefore we get the term $\frac{\tilde C}{N}$ where all unknown constants have been absorbed into $\tilde C$.
\end{enumerate}
Put together, this yields the following probability of alignment for the $N$th example:
\begin{align}
    P(N\text{th aligned}) = P(N\text{th already aligned}) \cdot 1 + (1 - P(N\text{th already aligned})) \cdot e^{-\frac{N-1}{A}}
    \\
    P(N\text{th already aligned}) = \frac C N \cdot \frac{1}{2} + \left(1 - \frac C N\right) \rho_{N-1}
\end{align}
Including the $N$th example, the fraction of correctly aligned updates is
\begin{equation}
    \rho_N = \frac{(N-1) \rho_{N-1} + P(N\text{th already aligned})}{N}.
\end{equation}
Taking $\rho$ as a continuous function, this expression can be interpreted as a differential equation by setting $\frac{\partial \rho(N)}{\partial N} = \rho_{N+1} - \rho_N$.
With the inital condition $\rho(1) = 1$, it can be solved analytically but results in a cumbersome integral expression.

\clearpage

\section{Baselines and network architectures} \label{sec:app:baselines-nets}

This section details the neural network architectures we employ in our experiments, as well as implementation details of the baselines to which we compare JPO.

\paragraph{Network architectures.}

We deliberately use generic off-the-shelf neural network architectures.
In all performed experiments, the solution space consists of a finite number of scalars while the initial and final simulation states often involve spatial data, i.e. grids.
For grid data, we use convolutional layers on multiple resolution levels while fully-connected layers process non-grid data.
Applying this approach to our problems results in the following scenarios:

\begin{itemize}
    \item \textbf{Grid to scalars (G2S).}
    We use a standard architecture for classification that largely follows by VGG~\citep{VGG2014}. It consists of multiple convolutional blocks followed by fully-connected layers.
    Each convolutional block consists of one or multiple convolutional layers with kernel size $3^d$ where $d$ denotes the number of grid dimensions.
    A batch normalization, activation, and max pooling layer follows each convolutional layer, reducing the resolution by half at the end of each block.
    The result is passed to a multilayer perceptron (MLP) which alternates linear, activation, and batch normalization layers before the result is outputted by a final linear layer.
    \item \textbf{Scalars to scalars (S2S).} When no grid data is involved, we simply use MLPs~\citep{DeepLearningBook} to map inputs to outputs, optionally with positional encoding at the inputs.
    The MLP consists of multiple linear layers and activation layers but we do not use batch normalization layers since our networks are relatively shallow with no more than three hidden layers.
    \item \textbf{Grid to grid (G2G).} This case is only needed for the surrogate network required by the neural adjoint method. Here, we use the popular U-Net~\citep{UNet} architecture with residual connections.
    Multiple convolutional blocks progressively decrease the spatial resolution, followed by upsampling convolutional blocks.
    The downsampling blocks match the ones described above. The upsampling blocks linearly interpolate the result to double the resolution before concatenating the corresponding processed input of the same resolution for the residual connections.
\end{itemize}

The specific hyperparameter values used for these generic architectures are given in the corresponding experimental details sections.


\paragraph{BFGS.}
We use the BFGS Implementation from SciPy~\citep{SciPy2020} which runs the optimizer-internal computations on the CPU.
All loss and gradient evaluations are bundled and dispatched to the GPU to be processed in parallel.

\paragraph{Supervised training}
Our main goal is obtaining an optimization scheme that works exactly like classical optimizers, only requiring the forward process $F$, $\gamma_i$ in the form of a numerical simulator, and desired outputs $y_i$. 
%
However, if we additionally have a prior on the solution space $P(\xi)$, we can generate synthetic training data $\{(x_j, y_i), \xi_j\}$ with $y_j = F(x_j, \xi_j)$ by sampling $x_i \sim P(\xi)$.
Using this data set, we can alternatively train $\mathcal N$ with the supervised objective
\begin{equation}
    \tilde L = \sum_j || \mathcal N(x_j, y_j) - \xi_j ||_2^2.
\end{equation}
Since $\mathcal N$ has the same inputs and outputs, we can use the same network architecture as above and the solutions to the original inverse problems can be obtained as $x_i = \mathcal N(\gamma_i, y_i)$.
While this method requires domain knowledge in the form of the distributions $P(x)$ and $P(\xi)$, it has the distinct advantage of being independent of the characteristics of $F$.
For example, if $F$ is chaotic, directly optimizing through $F$ can yield very large and unstable gradients, while the loss landscape of $\tilde L$ can still be smooth.
However, we cannot expect the inferred solutions to be highly accurate as the network is not trained on the inverse problems we want to solve and, thus, has to interpolate.
Additionally, this method is only suited to unimodal problems, i.e. inverse problems with a unique global minimum.
On multimodal problems, the network cannot be prevented from learning an interpolation of possible solutions, which may result in poor accuracy.

\paragraph{Neural adjoint}
The neural adjoint method~\citep{NeuralAdjoint} employs a neural network $\tilde{\mathcal N}$ to act as a surrogate for $F$.
$\tilde{\mathcal N}$ is then used in place of $F$ in an iterative optimization.
Since $\tilde{\mathcal N}$ cannot be expected to produce accurate results outside of the region covered by its training data, a boundary loss term is added to the optimization to prevent the optimizer from leaving that region~\citep{NeuralAdjoint}.
We formulate this boundary loss in a differentiable manner to make it compatible with higher-order optimizers.
First, we determine the minimum $\xi^j_\mathrm{min}$ and the maximum $\xi^j_\mathrm{max}$ value in the training set for each parameter $j$.
Then, we formulate the boundary loss as
\begin{align*}
    B(\xi) = \sum_j \mathrm{SoftPlus}_\gamma\left(\frac{\max(\xi^j - \xi^j_\mathrm{max}, \xi^j_\mathrm{min} - \xi^j)}{\xi^j_\mathrm{max} - \xi^j_\mathrm{min}} \right),
\end{align*}
where $\mathrm{SoftPlus}_\gamma(x) \equiv \frac 1 \gamma \log(1 + e^{\gamma x})$.

\newpage
\section{Experimental details} \label{app:experiments}

This section lists additional details about the experiments showcased in the main paper.
An overview of the symbols introduced with the experiments is given in Tab.~\ref{tab:app:symbols}.
By \emph{refinement}, we refer to a secondary optimization stage using BFGS.
The best neural network estimate $x_i$ for each solution is used as an initial guess for the BFGS optimizer which directly optimizes the individual objective function $L_i$ until convergence.

\begin{table}[htb]
\centering
\caption{Physical quantities corresponding to the abstract symbols used in Eqs.~1 and 2 for each experiment.}
\label{tab:app:symbols}
\begin{tabular}{llll}
\toprule
Experiment             & $\xi$           & $x$                    & $y$                  \\
\midrule
Wave packet fit        & $t_0$           & $\epsilon(t)$          & $u(t)$               \\
Billiards              & $\vec v_0$      & Initial ball positions & Final ball positions \\
Kuramoto–Sivashinsky   & $\alpha,\beta$  & $u(x)|_{t=0}$          & $u(x)|_{t=25}$       \\
Incompr. Navier-Stokes & $x_0, \vec v_0$ & $u(x,y)|_{t=0}$        & $u(x,y)|_{t=56}$     \\
\bottomrule
\end{tabular}
\end{table}

\paragraph{Software and hardware.}
We used PyTorch~\citep{PyTorch2019} and $\Phi_\textrm{Flow}$~\citep{phiflow} to run our experiments.
The full source code to reproduce our experiments will be made public under the MIT license upon publication.
The first three experiments were run on a GeForce RTX 3090. Due to memory requirements, the fluid experiment was run partly on a Quadro RTX 8000 which allowed 128 simulations to be held in GPU memory.

The corresponding wall-clock run times are shown in Tab.~\ref{tab:times}.
There, \emph{parallel BFGS} denotes a BFGS implementation that runs the forward process and backpropagation on the GPU.
This is much more efficient than looping over the individual examples, the way most classical BFGS optimizations are implemented.
For the KS experiment, sequential BFGS solves take 8x longer than the batched solve for $n=16$ and 76x longer for $n=256$.
Running this on the CPU increases the runtime by an additional 50-60\%.
Compared to the sequential CPU approach at $n=256$, our method is 18x faster than BFGS.

\begin{table}[h]
\caption{Training and optimization times for the largest tested data set size $n$ in seconds.}
\resizebox{\textwidth}{!}{%
\begin{tabular}{lllllllll}
\toprule
Experiment             & Parallel BFGS    & Network fit     & Refinement      & Supervised fit & Sup. Refinement & Surrogate fit   & Neural Adjoint & N.A. Refinement  \\
\midrule
Wave packet fit        & $15.1 \pm 1.0$   & $46.9 \pm 0.2$  & $15.0 \pm 0.2$  & $24.0 \pm 0.2$ & $13.8 \pm 0.7$        & $46.5 \pm 0.2$  & $13.1 \pm 0.5$ & $13.8 \pm 1.7$   \\
Billiards              & $21.5 \pm 1.0$   & $115.2 \pm 0.6$ & $25.3 \pm 0.2$  & $8.8 \pm 0.1$  & $20.9 \pm 1.7$        & $12.9 \pm 0.1$  & $16.5 \pm 2.5$ & $21.8 \pm 1.1$   \\
Kuramoto–Sivashinsky   & $152.8 \pm 11.8$ & $638.8 \pm 3.7$ & $109.3 \pm 7.2$ & $11.4 \pm 0.9$ & $122 \pm 64$      & $16.4 \pm 0.8$  & $13.7 \pm 2.3$ & $147 \pm 12$ \\
Incompr. Navier-Stokes & $1858 \pm 95$    & $29510 \pm 637$ & $1270 \pm 205$  & $212 \pm 7$    & $1390 \pm 333$        & $195.6 \pm 0.1$ & $8.7 \pm 1.2$  & $1451 \pm 63$   \\
\bottomrule
\end{tabular}
}
\label{tab:times}
\end{table}

\paragraph{Hyperparameter selection.}
For each network, we select one of the three generic architectures listed above.
Our main objective in choosing the values of the hyperparameters, such as the number of layers and layer width, is keeping the total number of parameters large enough to fit the problem easily but low enough to train the network quickly.
The only hyperparameter which we tune is the Adam~\citep{Adam} learning rate $\eta$.
We start with $\eta = 0.01$ and progressively reduce it by a factor of 10 until the loss decreases during the optimization.
The exact hyperparameter values are given in the corresponding experiment section.

\paragraph{Refinement.}
We apply BFGS refinement as a second stage to JPO, supervised training, and the neural adjoint method.
In all cases, we run a standard BFGS optimization on the actual $\mathcal L$, so the gradients are backpropagated through $F$.
For JPO and the neural adjoint method, we use the parameter estimate with the lowest recorded loss value.
As supervised training makes use of pre-trained network, we use the final parameter estimate $\xi_i$ as an initial guess. 
The full evolution of the parameter estimates over the course of training is shown for all experiments below.

Tab.~\ref{tab:improvement-fractions} lists the fraction of the total loss improvement performed by the network fit. In all experiments, the network fit stage is responsible for the bulk of the improvement and the refinement stage improves the loss much less overall. We also observe that, for larger data set sizes, the network fit contributes even more to the overall improvement while for small data set sizes, the refinement stage is more important.

\paragraph{Computation of improvement fractions}
The last plot in Figs.~\ref{fig:wavepacket}-\ref{fig:fluid} shows the average fraction of examples for which a method finds better solutions than BFGS.
Since some examples see multiple optimizers find the exact same solution, we add half of these cases to the shown values, i.e. the shown fractions $f_N$ are computed as
$$ f_N = \frac{N_\text{better}}{N} + \frac{N_\text{equal}}{2N}. $$
A value of $f_N=1$ occurs only if an optimizer finds better solutions than BFGS for all examples, $f_N=0$ means that all solutions are worse than BFGS, as expected.
We perform all experiments multiple times by sampling new data sets and initial network weights $\theta$.
Then we compute the mean and standard deviation over the so-obtained sets of $f_N$.
For $K$ repetitions, let $f_N^k$ denote the observed fraction for seed $k$.
Then, we plot the error bar as 
$$ \frac{\sqrt{\sum_k (f_N^k - \bar f_N)^2}}{N} \quad \text{with } \bar f_N = \frac{1}{M} \sum_k f_N^k . $$


\paragraph{Results.}
Tab.~\ref{tab:better-fractions} summarizes the improvements over classical optimizations for all methods with refinement and Tab.~\ref{tab:app:better-fractions} \emph{without} refinement.
Learning curves, loss and improvement statistics, as well as example parameter trajectories are shown in the following subsections.

\begin{table}[hb]
\caption{\label{tab:improvement-fractions} Fraction of the total loss decrease achieved by the network fit. The remaining improvement is made by the refinement stage using BFGS. The given fractions are computed per example and then averaged.}
\begin{tabular}{lllll}
\toprule
Experiment             & $n=4$                 & $n=8$                 & $n=32$                & $n=128$              \\ \midrule
Wave packet fit        & $78.5\% \pm 17.8\%$ & $89.1\% \pm 8.8\%$  & $92.4\% \pm 3.7\%$  & $91.7\% \pm 4.5\%$ \\ 
Billiards              & $88.9\% \pm 13.0\%$ & $86.8\% \pm 14.0\%$ & $92.9\% \pm 11.2\%$ & $98.1\% \pm 2.0\%$ \\
Kuramoto–Sivashinsky   & $93.4\% \pm 9.8\%$  & $96.2\% \pm 5.9\%$  & $96.0\% \pm 2.5\%$  & $95.9\% \pm 1.1\%$ \\
Incompr. Navier-Stokes & $100.0\% \pm 0.0\%$ & $99.4\% \pm 0.5\%$  & $96.6\% \pm 3.4\%$  & $96.8\% \pm 2.5\%$ \\
\bottomrule
\end{tabular}
\end{table}

\begin{table}[hb]
\centering
\caption{Fraction of inverse problems for which neural-network-based methods find better or equal solutions than BFGS. The best estimate of the network is selected for each inverse problem. Mean over multiple seeds and all $n$ shown in subfigures (d) of the main paper.}
\label{tab:app:better-fractions}
\begin{tabular}{lllllll}
\toprule
Experiment             & \multicolumn{2}{l}{JPO} & \multicolumn{2}{l}{Supervised} & \multicolumn{2}{l}{Neural Adjoint} \\
                       & Better            & Equal           & Better         & Equal         & Better           & Equal           \\
\midrule
Wave packet fit        &  \textbf{80.5\%}           & 0.9\%           & 55.9\%         & 2.0\%         & 28.2\%           & 0.8\%           \\
Billiards              &  \textbf{44.3\%}           & 9.0\%           & 14.6\%         & 19.9\%        & 1.6\%            & 29.3\%          \\
Kuramoto–Sivashinsky   &  \textbf{42.8\%}           & 0.0\%           & 14.4\%         & 0.0\%         & 6.4\%            & 0.0\%           \\
Incompr. Navier-Stokes &  \textbf{62.5\%}           & 0.0\%           & 23.5\%         & 0.0\%         & 1.1\%            & 0.0\%           \\
\bottomrule
\end{tabular}
\end{table}

\begin{table}[hb]
\centering
\caption{Fraction of inverse problems for which neural-network-based methods \emph{with refinement}, i.e. with secondary BFGS optimization, find better or equal solutions than BFGS. Mean over multiple seeds and all $N$ shown in subfigures (d).}
\label{tab:better-fractions}
\begin{tabular}{lllllll}
\toprule
Experiment             & \multicolumn{2}{l}{JPO} & \multicolumn{2}{l}{Supervised} & \multicolumn{2}{l}{Neural Adjoint} \\
                       & Better            & Equal           & Better         & Equal         & Better           & Equal           \\
\midrule
Wave packet fit       & \textbf{86.0\%}   & 1.8\%           & 65.1\%         & 14.4\%        & 40.2\%           & 47.4\%          \\
Billiards              & \textbf{61.7\%}   & 9.0\%           & 27.0\%         & 27.2\%        & 1.6\%            & 98.4\%           \\
Kuramoto–Sivashinsky   & \textbf{62.3\%}   & 0.0\%           & 57.7\%         & 0.0\%         & 23.9\%           & 62.2\%          \\
Incompr. Navier-Stokes & 64.1\%            & 0.0\%           & \textbf{66.2\%}& 0.1\%         & 56.9\%           & 0.1\%           \\
\bottomrule
\end{tabular}
\end{table}

\clearpage
\subsection{Wave packet localization} \label{app:wavepacket}

For the wave packet experiment, we first determine the true position $t_0$ of the wave packet by sampling random values from a uniform distribution between $t_0 \in [26, 230]$.
Noise $\epsilon(t)$ is sampled from a normal distribution with standard deviation $\sigma=0.1$ for every $t = 1, ..., 256$ and superimposed on the signal, as described in the main text.
The noise pattern is only used to generate the reference data and is not available to the optimizers.
We run this experiment five times with varying initialization seeds for both networks and data sets.

\begin{figure}[htb]
    \centering
    \includegraphics[width=.8\textwidth]{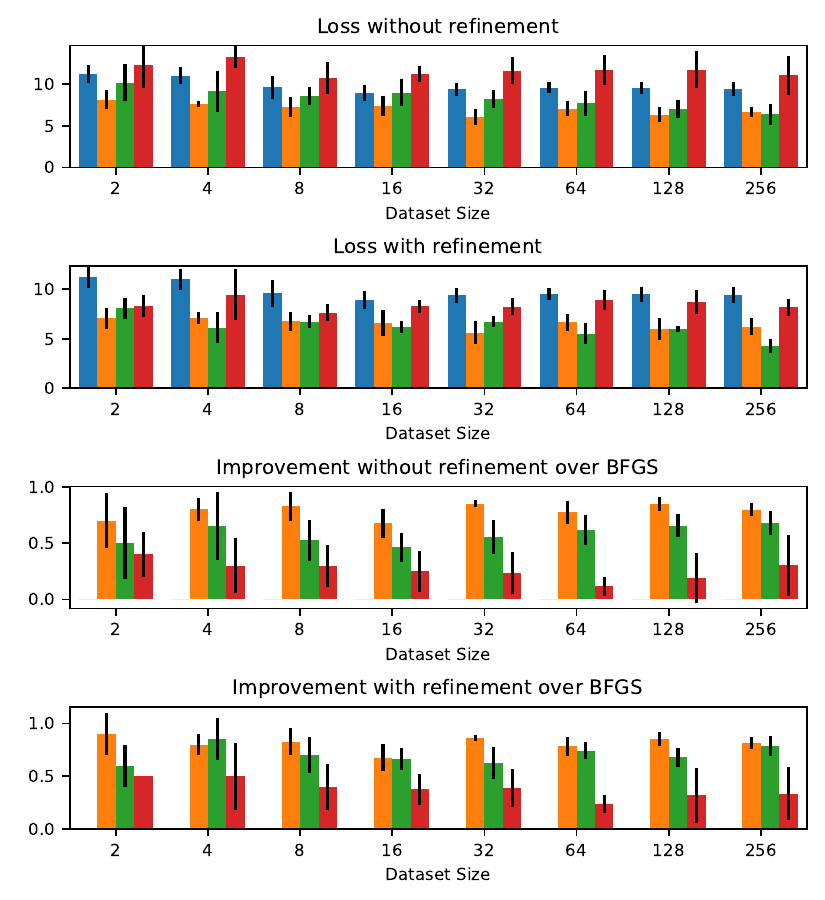}
    \caption{Loss and improvement over BFGS before and after refinement for the wave packet experiment.
    Colors match figures from the main paper (blue: BFGS, orange: JPO, green: supervised, red: neural adjoint).}
    \label{fig:app:wavepacket-improvement}
\end{figure}

\begin{figure}[htb]
    \centering
    \includegraphics[width=.8\textwidth]{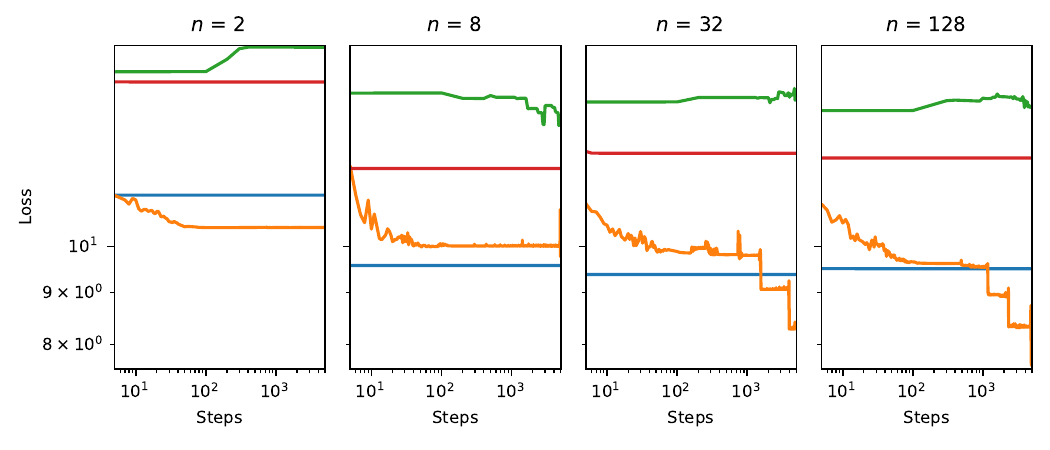}
    \caption{Optimization curves for different data set sizes of the wave packet experiment before refinement.
    Curves show the mean over 5 network and data set initialization seeds.
    Blue: BFGS, orange: JPO, green: supervised, red: neural adjoint.}
    \label{fig:app:wavepacket-curves}
\end{figure}

\begin{figure}[htb]
    \centering
    \includegraphics[width=.8\textwidth]{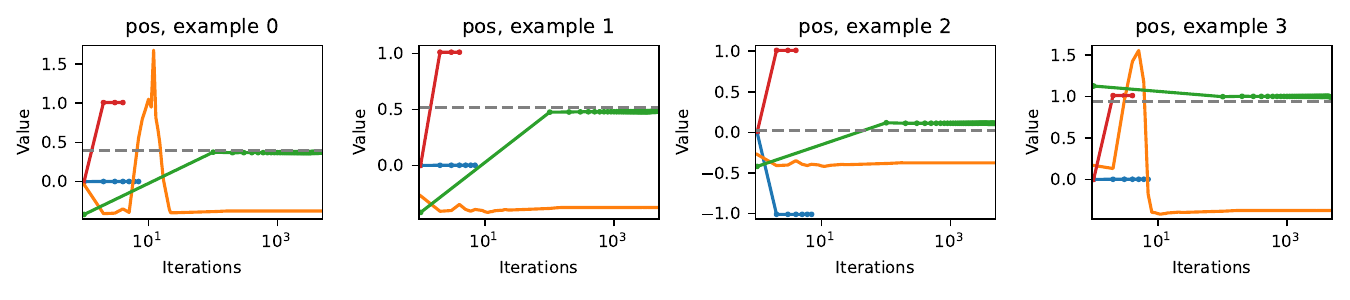}
    \caption{Example parameter evolution during optimization of the wave packet experiment with $n=128$.
    Blue: BFGS, orange: JPO, green: supervised, red: neural adjoint.
    The dashed gray lines indicate the reference solution from which the example was generated.
    BFGS-based optimization curves stop when all examples have fully converged to an optimum.}
    \label{fig:app:wavepacket-params}
\end{figure}

\begin{figure}[htb]
    \centering
    \includegraphics[width=.8\textwidth]{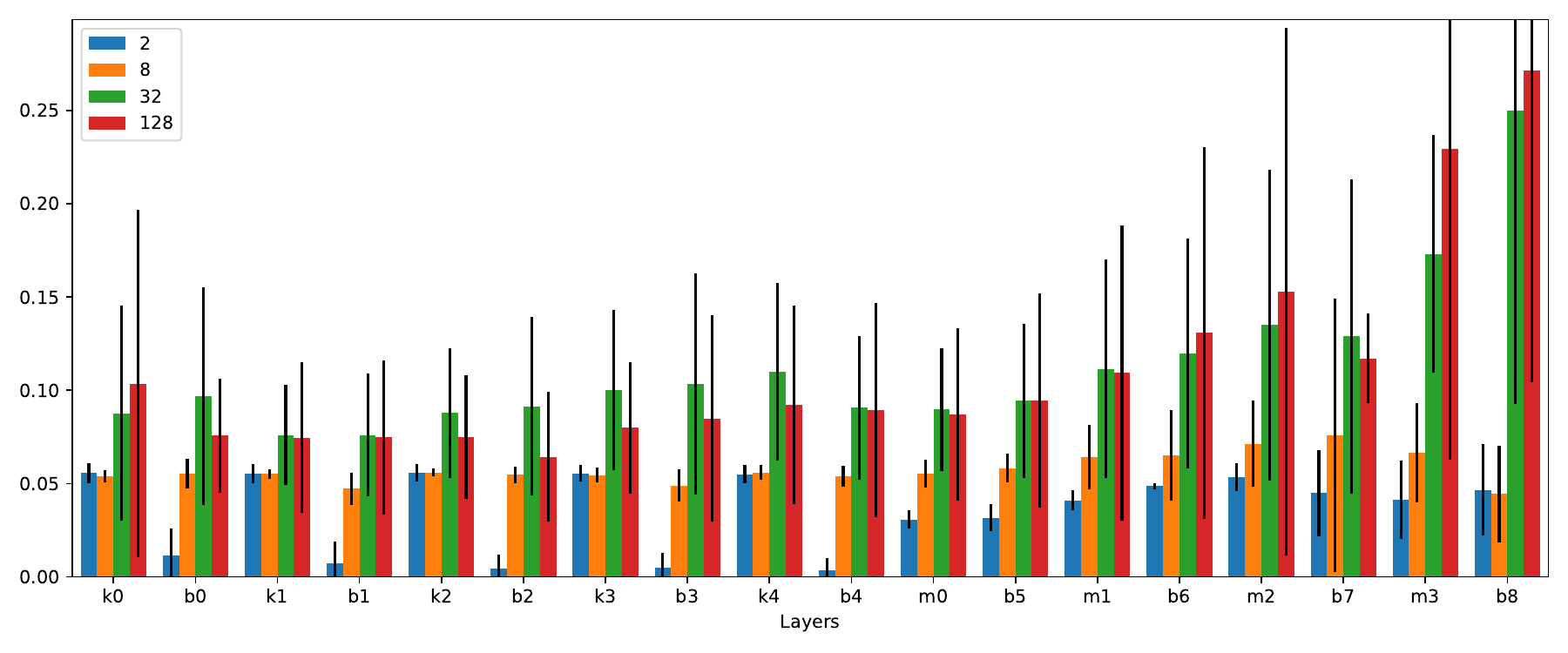}
    \caption{JPO network change in the wave packet experiment for different data set sizes $n$, measured as the mean absolute difference in weight values before and after fitting the network. Error bars represent the standard deviation across multiple network initializations. The change is given per layer where k denotes convolution kernels, m the matrices of fully-connected layers and b biases.}
    \label{fig:app:wavepacket-weight-change}
\end{figure}

\paragraph{Networks.}
The surrogate network, required by the neural adjoint method, takes $t_0$ and $\epsilon(t)$ as input and outputs an approximation of $u(t)$.
Since $\epsilon(t)$ and $u(t)$ are one-dimensional grids, we employ the G2G architecture with two input feature maps of size 256 and one output feature map, totaling 13,073 parameters.
The JPO network maps the grid $u(t)$ to the estimated scalar $t_0$.
Consequently, we use the G2S architecture described in section~\ref{sec:app:baselines-nets}.
We use five blocks with one convolutional layer with 16 feature maps each, reducing the resolution from 256 to 8.
The MLP part consists of two hidden fully-connected layers with sizes 64 and 32.
In total, this network contains 13,925 parameters.
All networks are trained using Adam with a learning rate of $\eta = 0.001$.
Fig.~\ref{fig:app:wavepacket-weight-change} shows the absolute the change in the JPO network weights that results from training.

\paragraph{Additional results.}
Fig.~\ref{fig:app:wavepacket-improvement} shows the resulting loss and improvement over BFGS, both before and after refinement.
The learning curves for four data set sizes $n$ are shown in Fig.~\ref{fig:app:wavepacket-curves},
and the parameter evolution of four examples during optimization are shown in Fig.~\ref{fig:app:wavepacket-params}.
The neural adjoint method finds better solutions than BFGS for about a third of examples for $N=256$ (see Tab.~\ref{tab:better-fractions}).
In many cases, the optimization progresses towards the boundary and gets stuck once the boundary loss $B$ balances the gradients from the surrogate network.
For supervised training of $\mathcal N$, we use the same training data set as for the neural adjoint method.
This approach's much smoother loss landscape lets all solution estimates progress close to the ground truth.
However, lacking the gradient feedback from the forward process $\mathcal F$, the inferred solutions are slightly off from the actual solution and, since the highest loss values are close to the global optimum, this raises the overall loss during training even though the solutions are approaching the global optima.

\clearpage
\subsection{Billiards}
For the billiards experiment, we set up a rigid body simulation of spherical balls with radius $r=0.2$ moving in the x-y-plane.
In each step, the simulator analytically integrates the evolution until the time of the subsequent collision, allowing us to simulate the dynamics at little computational cost. 
Collisions use a fixed elasticity of 0.8 and preserve momentum.
Friction is assumed to be proportional to the speed of the balls.
The simulation stops once no more collisions take place and integrates up to $t=\infty$ to let all balls come to rest.
we sample initial states by randomly placing the second ball between $(1, 0)$ and $(1, 1)$ while keeping the target fixed at $(2, 0.5)$.
The cue ball, located at $x=0$ is given a starting initial velocity of $\vec v_0^\mathrm{start} = (1, 0)$ so it will collide with the second ball in many cases by default.
Starting the optimization with $\vec v_0^\mathrm{start} = 0$ would yield $\nabla L_i = 0 \, \forall i$ and prevent any optimization using $F$.
However, in none of these examples does the ball exactly reach the target.
As the distribution of actual solutions $\vec v_0$ is unknown, the generated training sets for supervised and surrogate network training must rely on this broader data set, making learning more difficult.

In this experiment, the distribution of the solutions $P(\vec v_0)$ is not available as hitting the target precisely requires a specific velocity $\vec v_0$ that is unknown a-priori.
We can, however, generate training data with varying $\vec v_0$ and observe the final positions of the balls, then train a supervised $\mathcal N$ as well as a surrogate network for the neural adjoint method on this data set.
However, this is less efficient as most of the examples in the data set do not result in an optimal collision.

\begin{figure}[htb]
    \centering
    \includegraphics[width=.8\textwidth]{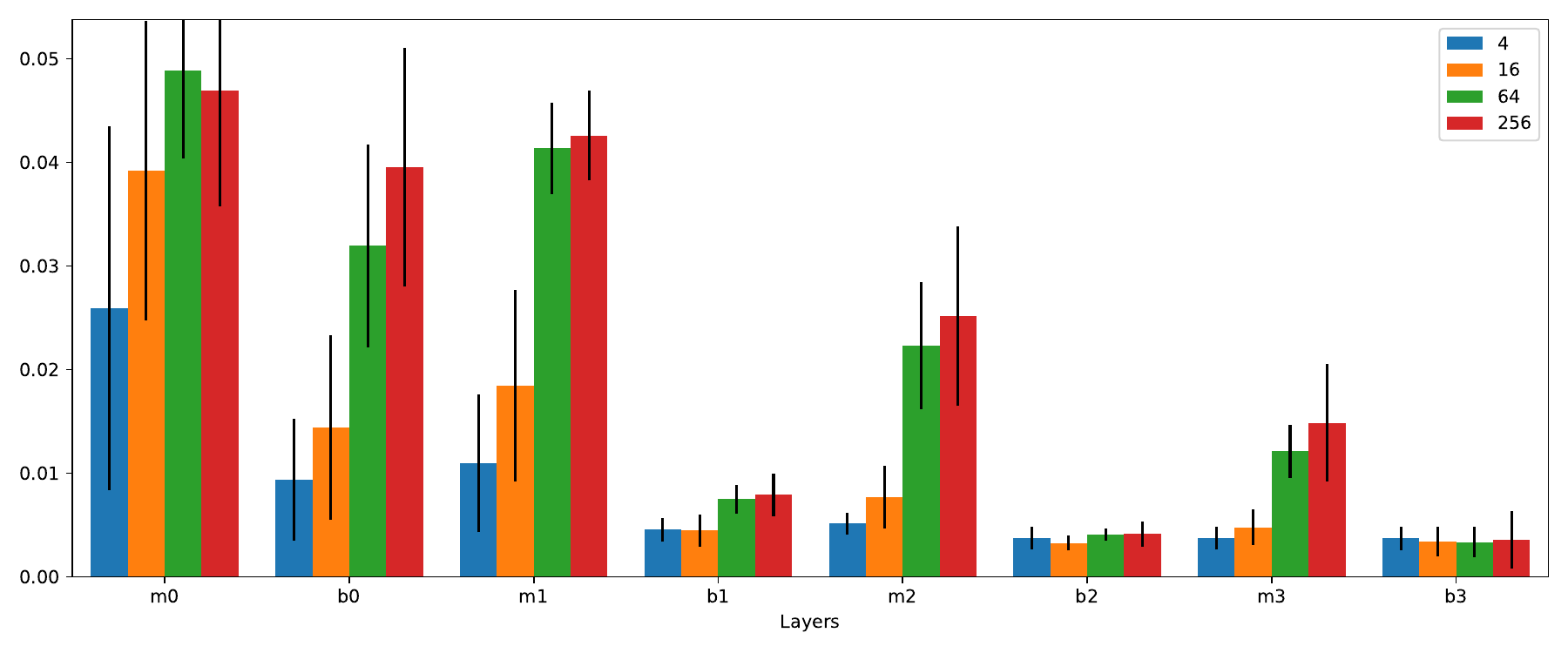}
    \caption{JPO network change in the Billiards experiment for different data set sizes $n$, measured as the mean absolute difference in weight values before and after fitting the network. Error bars represent the standard deviation across multiple network initializations. The change is given per layer where k denotes convolution kernels, m the matrices of fully-connected layers and b biases.}
    \label{fig:app:billiards-weight-change}
\end{figure}

\paragraph{Networks.}
The surrogate neural network is given a value for the initial velocity $\vec v_0$ and the balls' positions as input, and it outputs the predicted final position of the ball.
We use positional encoding for the input using sine, cosine functions with four equidistant frequencies.
The surrogate network follows the S2S architecture from section~\ref{sec:app:baselines-nets}. It is an MLP with three hidden layers containing 128 neurons, each, and comprises 37,506 parameters in total.
The JPO network predicts $\vec v_0$ based on the initial and final ball positions, and we use the same network architecture as for the surrogate network.
All networks are trained using Adam.
For the JPO, we use a learning rate of $\eta = 10^{-4}$ while all other methods use $\eta = 0.001$.
Fig.~\ref{fig:app:billiards-weight-change} shows the absolute the change in the JPO network weights that results from training.

\begin{figure}[htb]
    \centering
    \includegraphics[width=.8\textwidth]{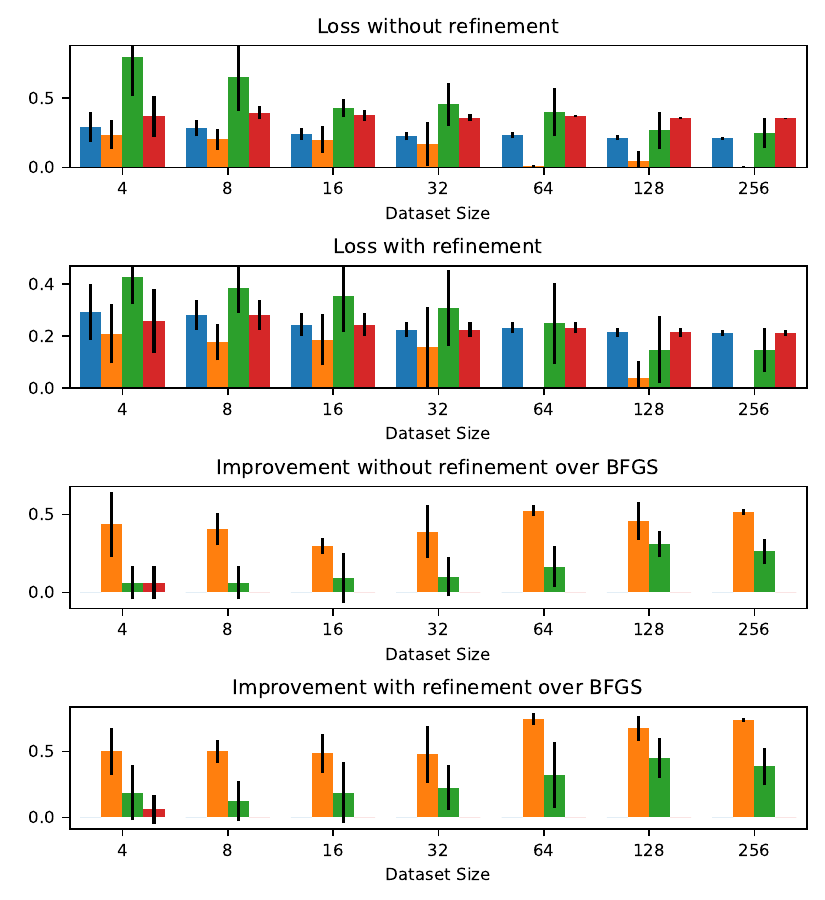}
    \caption{Loss and improvement over BFGS before and after refinement for the billiards experiment.
    Blue: BFGS, orange: JPO, green: supervised, red: neural adjoint.}
    \label{fig:app:billiards-improvement}
\end{figure}

\begin{figure}[htb]
    \centering
    \includegraphics[width=.8\textwidth]{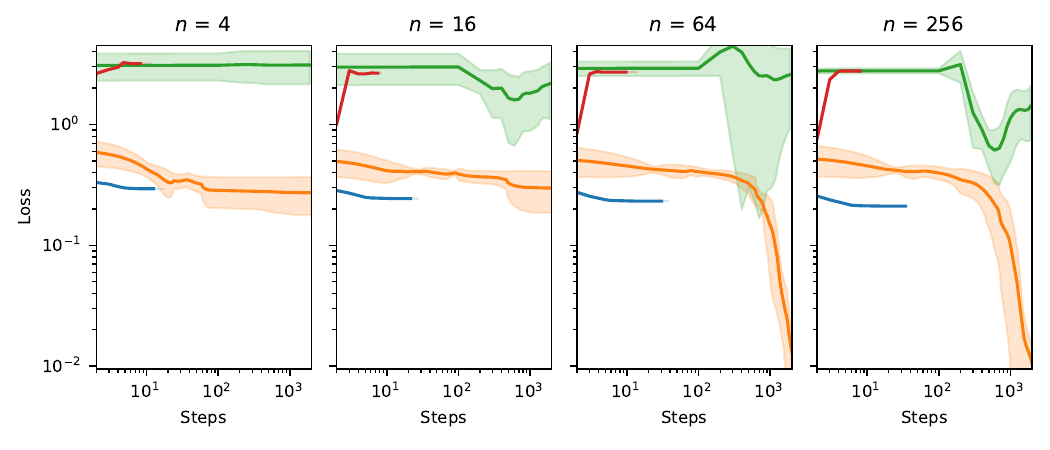}
    \caption{Optimization curves for different data set sizes of the billiards experiment before refinement.
    Envelopes show the standard deviation over 4 network and data set initialization seeds.
    Blue: BFGS, orange: JPO, green: supervised, red: neural adjoint.
    BFGS-based optimization curves stop when all examples have fully converged to an optimum.}
    \label{fig:app:billiards-curves}
\end{figure}

\begin{figure}[htb]
    \centering
    \includegraphics[width=.8\textwidth]{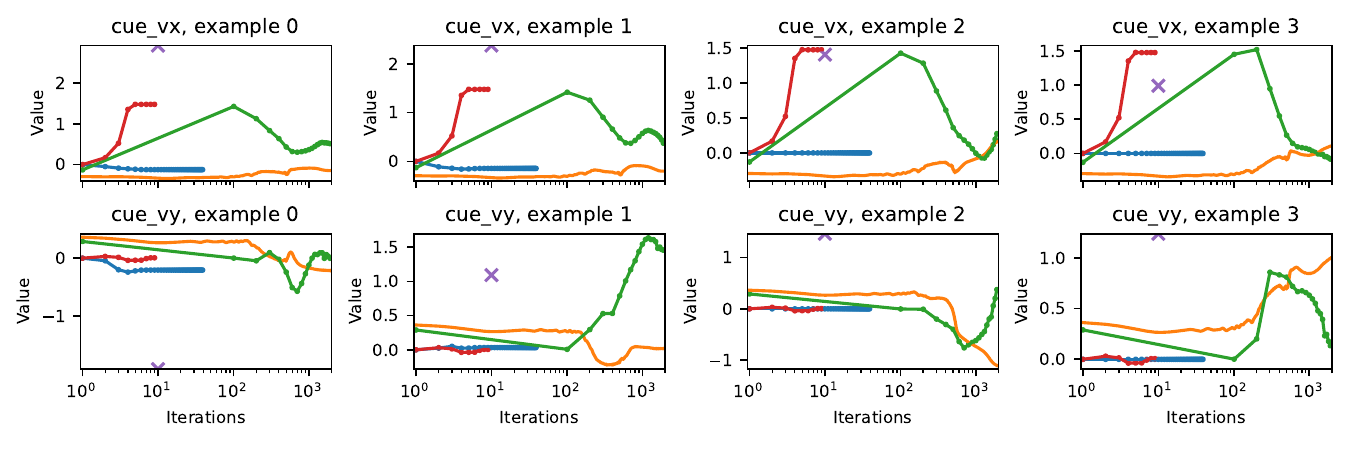}
    \caption{Example parameter evolution during optimization of the billiards experiment with $n=128$.
    Blue: BFGS, orange: JPO, green: supervised, red: neural adjoint.
    The purple crosses indicate the reference from which the example was generated and is not a valid solution in this experiment.
    BFGS-based optimization curves stop when all examples have fully converged to an optimum.}
    \label{fig:app:billiards-params}
\end{figure}

\paragraph{Additional results.}
Fig.~\ref{fig:app:billiards-improvement} shows the resulting loss and improvement over BFGS, both before and after refinement.
The learning curves for four data set sizes $n$ are shown in Fig.~\ref{fig:app:billiards-curves},
and the parameter evolution of four examples during optimization are shown in Fig.~\ref{fig:app:billiards-params}.
The neural adjoint method fails to approach the true solutions and instead gets stuck on the training data boundary in solution space.
Likewise, the supervised model cannot accurately extrapolate the true solution distribution from the sub-par training set.

\clearpage
\subsection{Kuramoto–Sivashinsky equation}
For this experiment, we set up a differentiable simulation of the Kuramoto–Sivashinsky (KS) equation in one dimension with a resolution of 128.
We simulate the linear terms of KS equation in frequency space and use a Runge-Kutta-2~\citep{NumericalRecipes} scheme for the non-linear term.
The initial state is sampled from random noise in frequency space with smoothing applied to suppress high frequencies.
In each simulation step, we add a forcing of the form $G(x) = 0.1 \cos(x) - 0.01 \cos(x/16) \cdot (1 - 2 \sin(x/16)$ which is controlled by the parameter $\alpha$ as described in the main text.

\begin{figure}[htb]
    \centering
    \includegraphics[width=.8\textwidth]{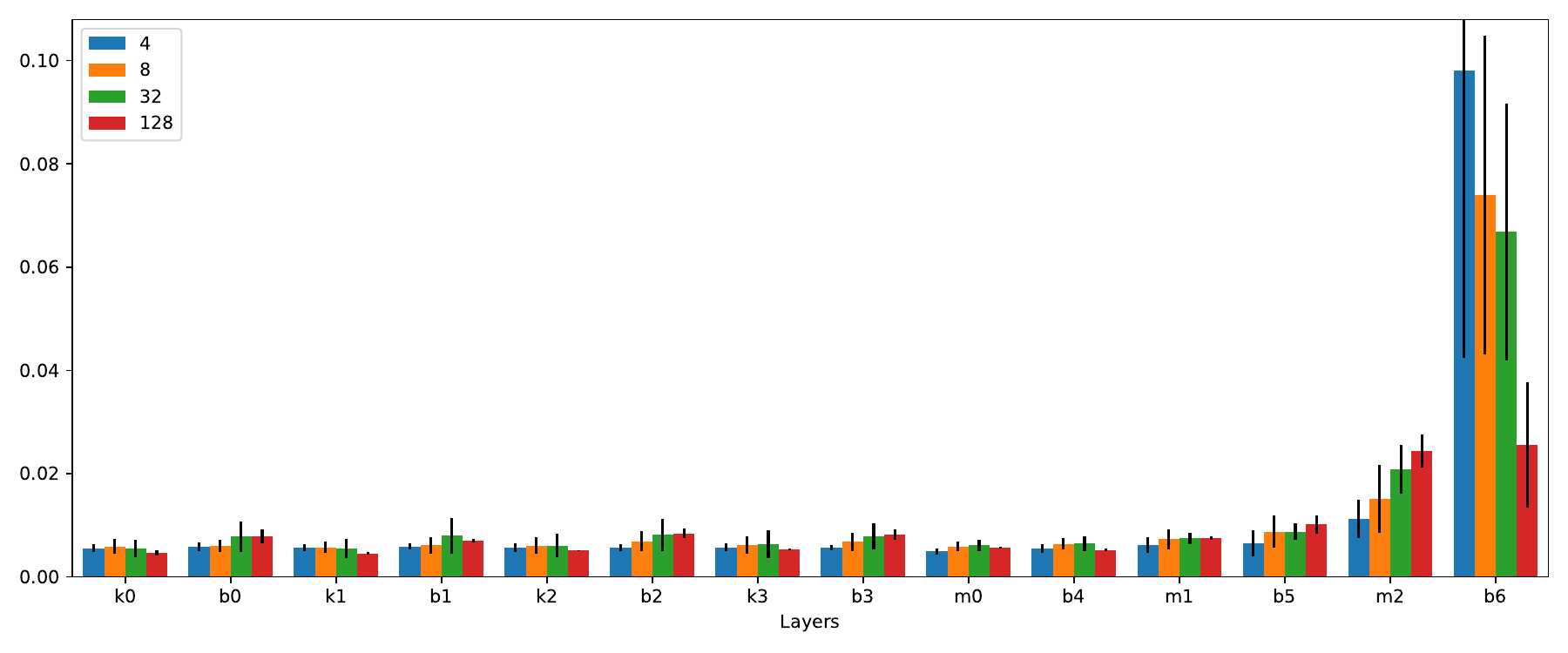}
    \caption{JPO network change in the Kuramoto–Sivashinsky experiment for different data set sizes $n$, measured as the mean absolute difference in weight values before and after fitting the network. Error bars represent the standard deviation across multiple network initializations. The change is given per layer where k denotes convolution kernels, m the matrices of fully-connected layers and b biases.}
    \label{fig:app:ks-weight-change}
\end{figure}

\paragraph{Networks.}
The surrogate network maps the initial state $u(x, t=0)$ and parameters $\alpha, \beta$ to the final state $u(x, t=25)$.
Since $u$ is sampled on a grid, we use the G2G architecture from section~\ref{sec:app:baselines-nets} with three input and one output feature map, operating on four resolution levels.
The JPO network maps $u(x, t=0)$ and $u(x, t=25)$ to $\alpha, \beta$ and we employ the G2S architecture with four convolutional layers of widths 32, 32, 64, 64, followed by two hidden fully-connected layers with 64 neurons each.
We train both networks using Adam with a learning rate of $\eta = 0.001$ for 1000 iterations.
Fig.~\ref{fig:app:ks-weight-change} shows the absolute the change in the JPO network weights that results from training.

\begin{figure}[htb]
    \centering
    \includegraphics[width=.8\textwidth]{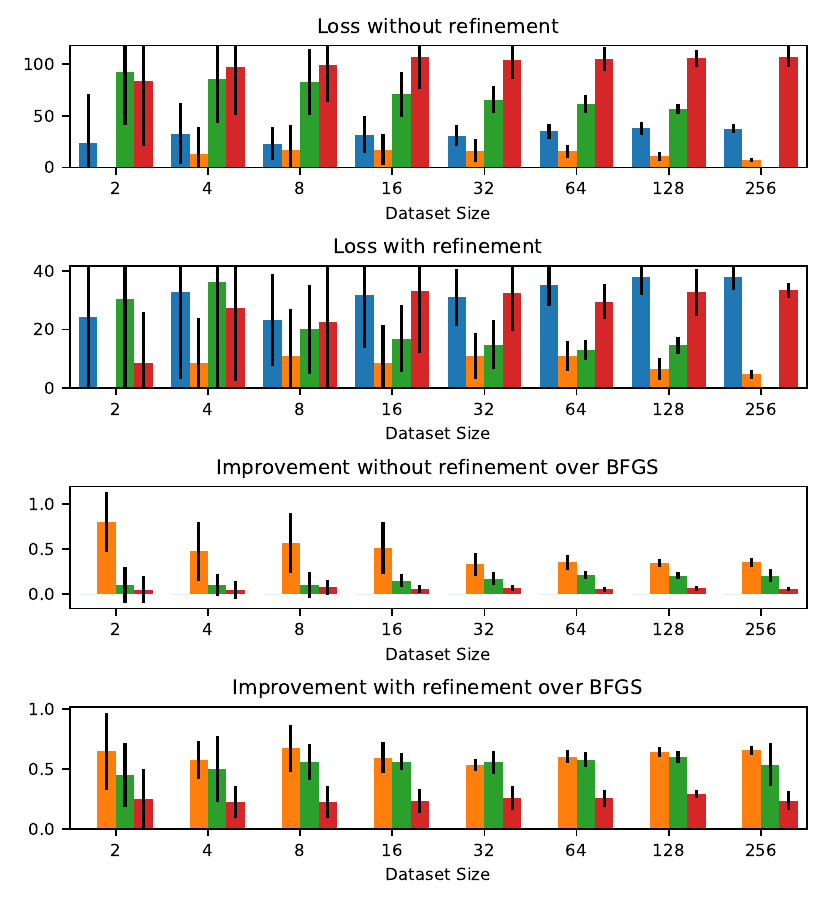}
    \caption{Loss and improvement over BFGS before and after refinement for the Kuramoto–Sivashinsky experiment.
    Blue: BFGS, orange: JPO, green: supervised, red: neural adjoint.}
    \label{fig:app:ks-improvement}
\end{figure}

\begin{figure}[htb]
    \centering
    \includegraphics[width=.8\textwidth]{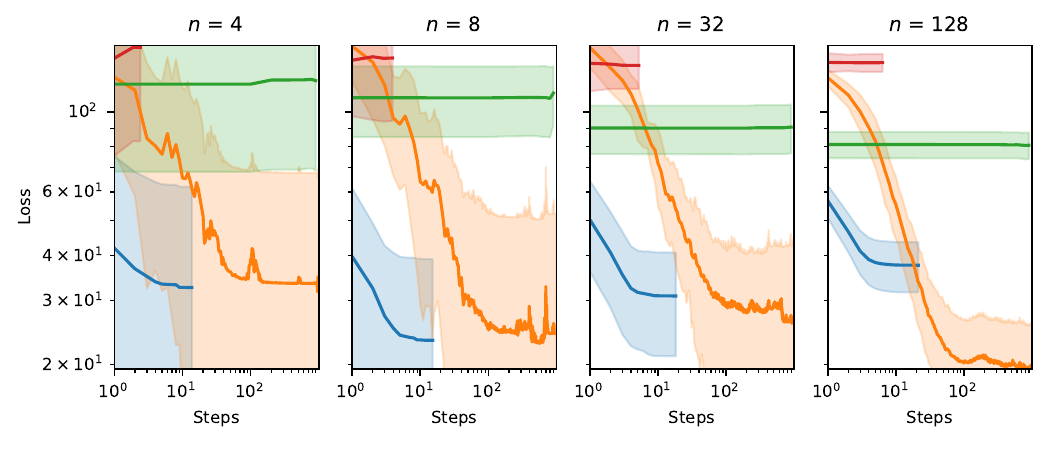}
    \caption{Optimization curves for different data set sizes of the Kuramoto–Sivashinsky experiment before refinement.
    Envelopes show the standard deviation over 10 network and data set initialization seeds.
    Blue: BFGS, orange: JPO, green: supervised, red: neural adjoint.
    BFGS-based optimization curves stop when all examples have fully converged to an optimum.}
    \label{fig:app:ks-curves}
\end{figure}

\begin{figure}[htb]
    \centering
    \includegraphics[width=.8\textwidth]{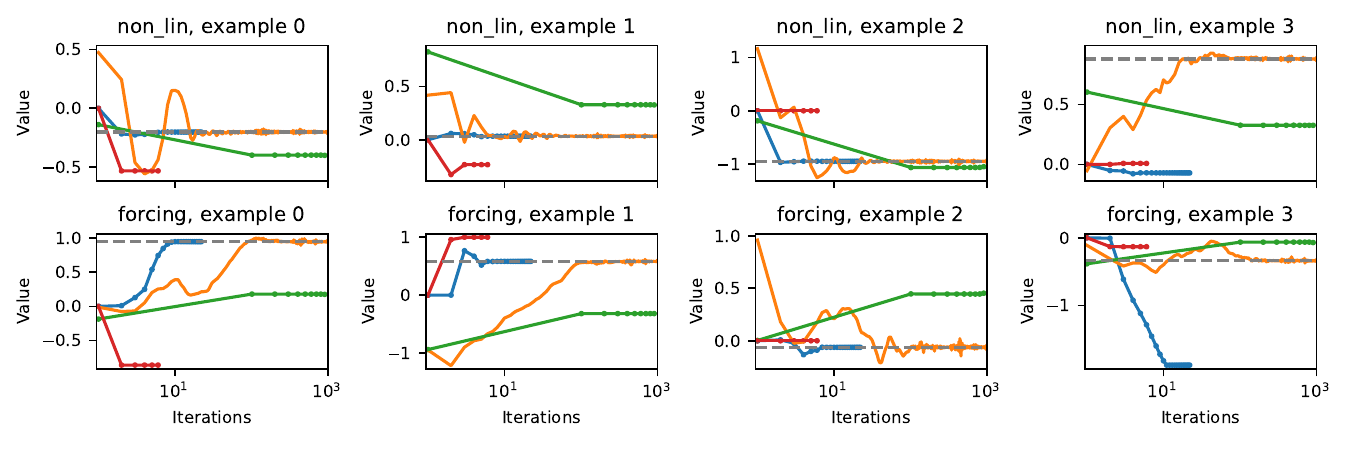}
    \caption{Example parameter evolution during optimization of the Kuramoto–Sivashinsky experiment with $n=128$.
    Blue: BFGS, orange: JPO, green: supervised, red: neural adjoint.
    The dashed gray lines indicate the reference solution from which the example was generated.
    BFGS-based optimization curves stop when all examples have fully converged to an optimum.}
    \label{fig:app:ks-params}
\end{figure}

\paragraph{Additional results.}
Fig.~\ref{fig:app:ks-improvement} shows the resulting loss and improvement over BFGS, both before and after refinement.
The learning curves for four data set sizes $n$ are shown in Fig.~\ref{fig:app:ks-curves},
and the parameter evolution of four examples during optimization are shown in Fig.~\ref{fig:app:ks-params}.

Supervised training with refinement produces better solutions in 58\% of examples, averaged over the shown $N$.
The unrefined solutions benefit from larger $N$ on this example because of the large number of possible observed outputs that the KS equation can produce for varying $\alpha,\beta$.
At $N=2$, all unrefined solutions are worse than BFGS while for $N \geq 64$ around 20\% of problems find better solutions.
With refinement, these number jump to 50\% and 62\%.

This property also makes it hard for a surrogate network, required by the neural adjoint method, to accurately approximate the KS equation, causing the following adjoint optimization to yield inaccurate results that fail to match BFGS even after refinement.

On a handful of examples, supervised learning and the neural adjoint method predicted parameter configurations for which the forward simulation diverged.
Due to their negligible impact on the results, we have dropped them from the diagrams.

\clearpage
\subsection{Incompressible Navier-Stokes}

\begin{figure}[bt]
    \centering
    \includegraphics[width=.8\textwidth]{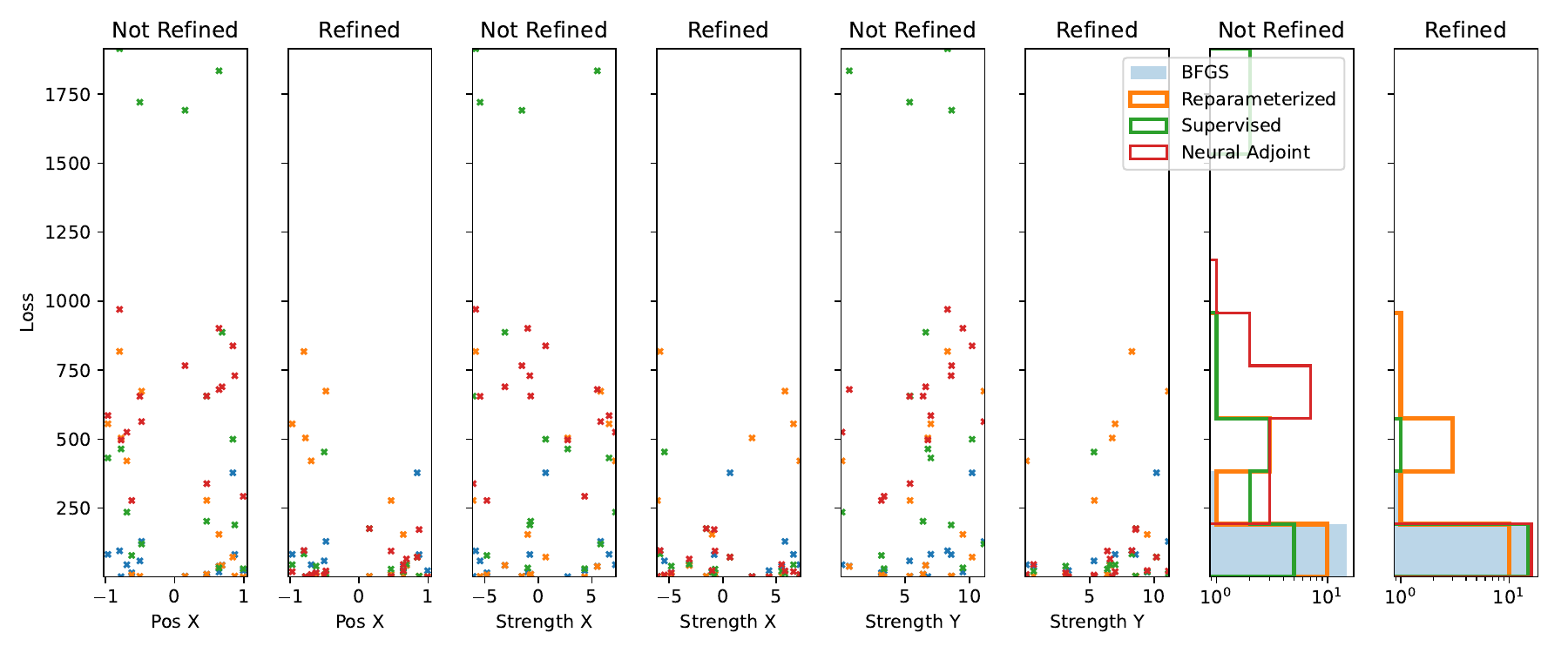}
    \caption{Distribution of loss values for $n=4$ in the Navier-Stokes experiment, with and without refinement. The first six plots show the loss distribution along the ground truth value of one of the parameters to be optimized. The right plots show the margin distribution of loss values. The results of 4 network and data set initialization seeds are accumulated.}
    \label{fig:app:fluid-loss-stat-4}
\end{figure}

\begin{figure}[bt]
    \centering
    \includegraphics[width=.8\textwidth]{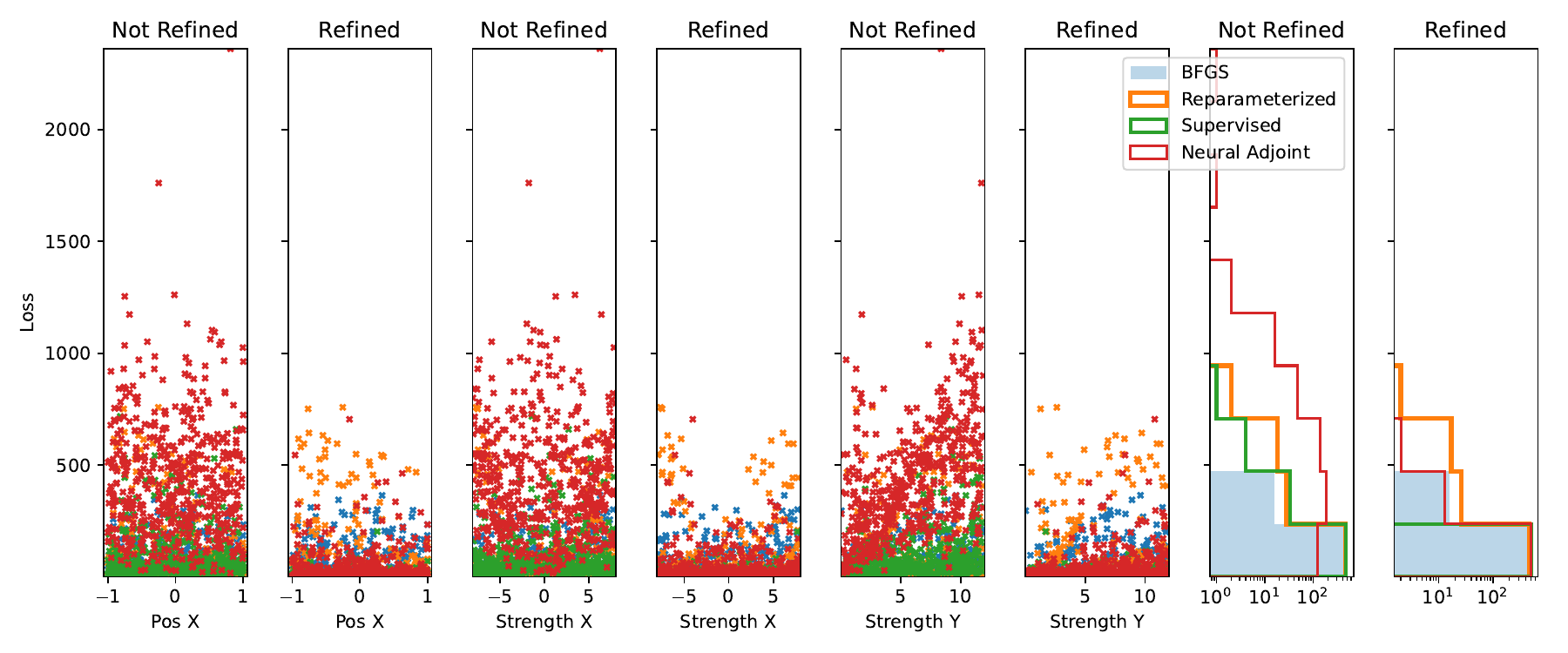}
    \caption{Distribution of loss values for $n=128$ in the Navier-Stokes experiment, with and without refinement. The first six plots show the loss distribution along the ground truth value of one of the parameters to be optimized. The right plots show the margin distribution of loss values. The results of 4 network and data set initialization seeds are accumulated.}
    \label{fig:app:fluid-loss-stat-128}
\end{figure}

\begin{figure}[htb]
    \centering
    \includegraphics[width=.8\textwidth]{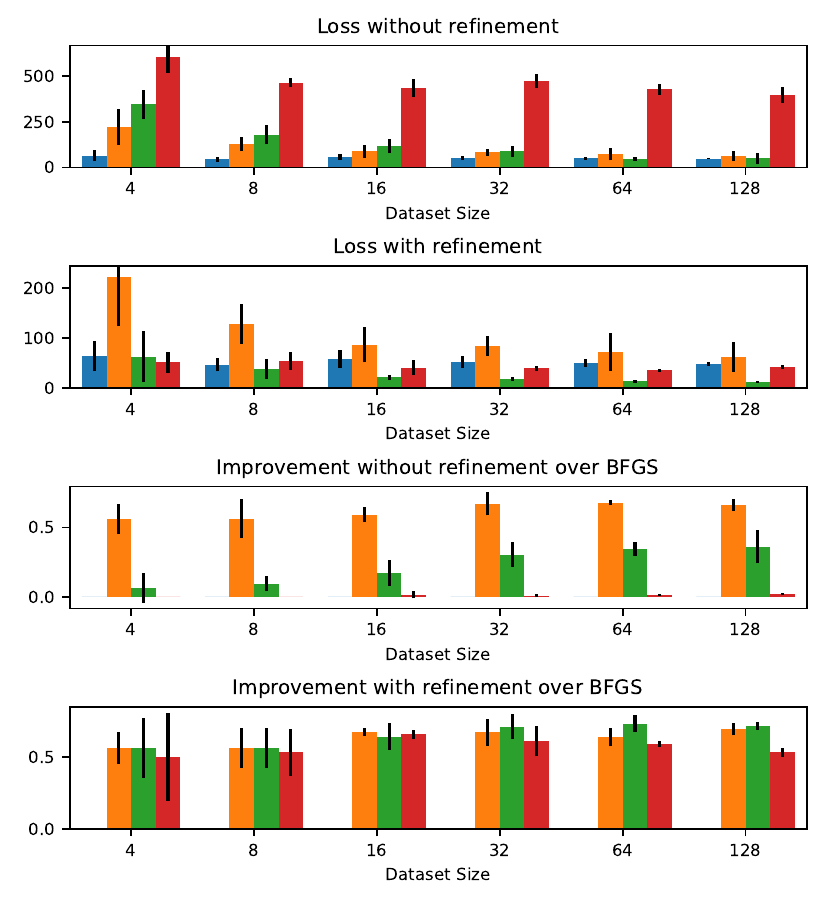}
    \caption{Loss and improvement over BFGS before and after refinement for the Navier-Stokes experiment.
    Blue: BFGS, orange: JPO, green: supervised, red: neural adjoint.}
    \label{fig:app:fluid-improvement}
\end{figure}

\begin{figure}[htb]
    \centering
    \includegraphics[width=.8\textwidth]{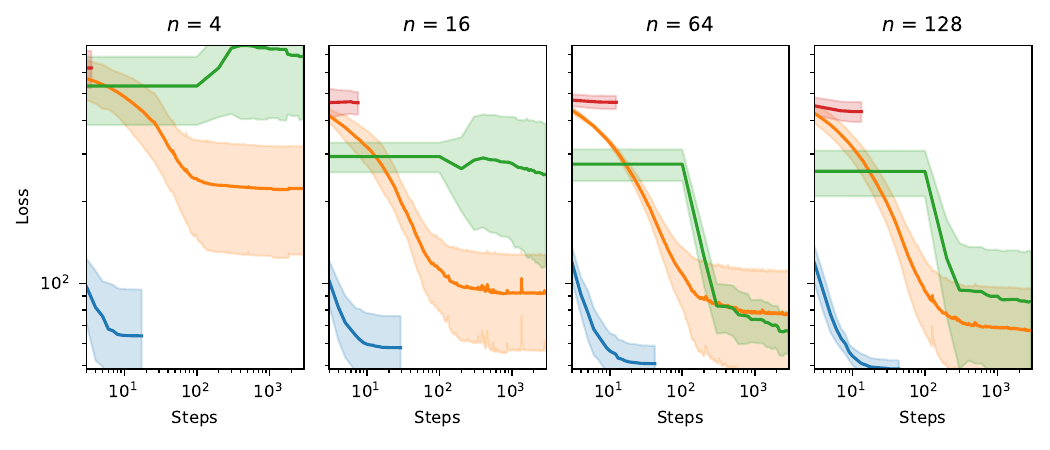}
    \caption{Optimization curves for different data set sizes of the Navier-Stokes experiment before refinement.
    Envelopes show the standard deviation over 4 network and data set initialization seeds.
    Blue: BFGS, orange: JPO, green: supervised, red: neural adjoint.
    BFGS-based optimization curves stop when all examples have fully converged to an optimum.}
    \label{fig:app:fluid-curves}
\end{figure}

\begin{figure}[htb]
    \centering
    \includegraphics[width=.8\textwidth]{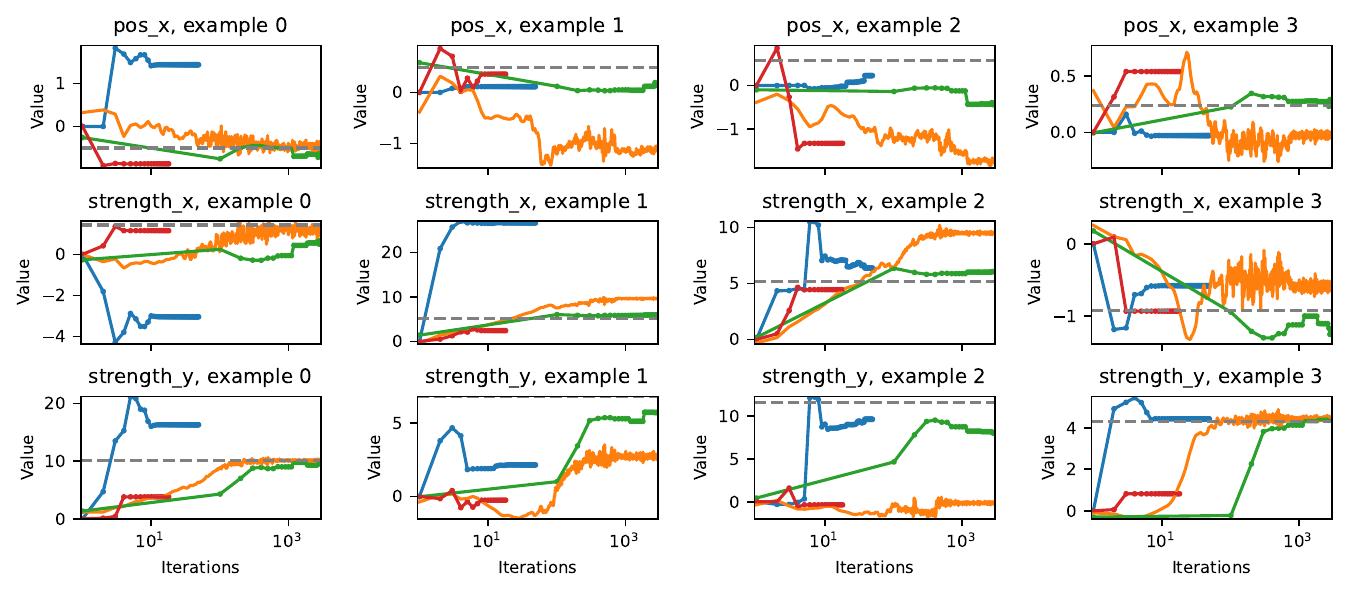}
    \caption{Example parameter evolution during optimization of the Navier-Stokes experiment with $n=128$.
    Blue: BFGS, orange: JPO, green: supervised, red: neural adjoint.
    The dashed gray lines indicate the reference solution from which the example was generated.
    BFGS-based optimization curves stop when all examples have fully converged to an optimum.}
    \label{fig:app:fluid-params}
\end{figure}

\begin{figure}[htb]
    \centering
    \includegraphics[width=.8\textwidth]{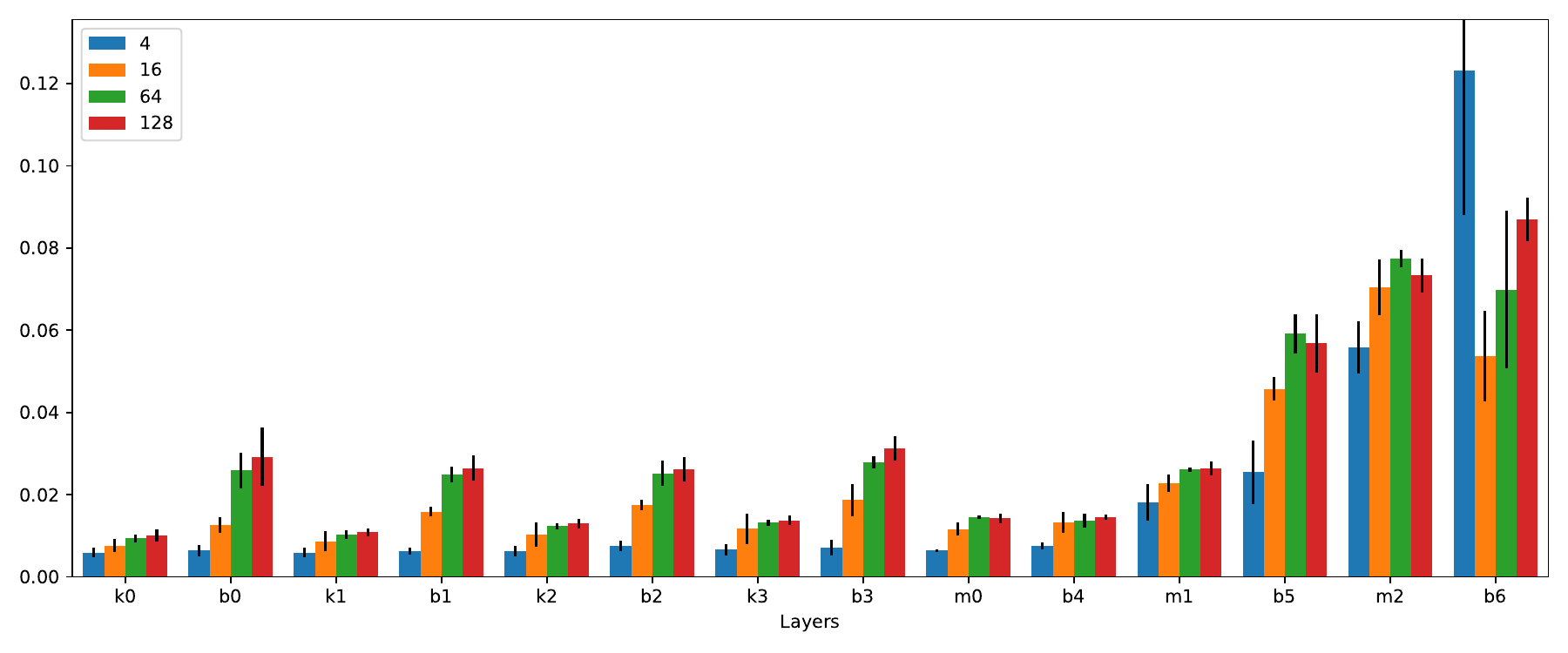}
    \caption{JPO network change in the incompressible fluid experiment for different data set sizes $n$, measured as the mean absolute difference in weight values before and after fitting the network. Error bars represent the standard deviation across multiple network initializations. The change is given per layer where k denotes convolution kernels, m the matrices of fully-connected layers and b biases.}
    \label{fig:app:fluid-weight-change}
\end{figure}

We simulate an incompressible two-dimensional fluid in a 100 by 100 box with a resolution of 64 by 64, employing a direct numerical solver for incompressible fluids from $\Phi_\textrm{Flow}$~\citep{phiflow}.
Specifically, we use the marker-in-cell (MAC) method~\citep{Mac1965, MAC} which guarantees stable simulations even for large velocities or time increments.
The velocity vectors are sampled in staggered form at the face centers of grid cells while the marker density is sampled at the cell centers.
The initial velocity $v_0$ is specified at cell centers and resampled to a staggered grid for the simulation.
Our simulation employs a second-order advection scheme~\citep{MacCormackStable} to transport the marker and the velocity vectors.
We do not simulate explicit diffusion as the numerical diffusion introduced by the advection scheme on this resolution is sufficient for our purposes.
Incompressibility is achieved via Helmholtz decomposition of the velocity field using a conjugate gradient solve.

We initialize the whole domain with a velocity field sampled from random noise in frequency space, resulting in eddies of various sizes.
The initial velocity values have a mean of zero and a standard deviation of 0.5.
Then, ground truth values for $x_0$ and $\vec v_0$ are sampled from uniform distributions with $\vec v_0^y \geq 0$ never pointing downward.
These values are used to initialize a spherical force or wind blast near the bottom of the domain that moves upwards during the simulation and induces flow around all obstacles from the pressure computation.
The velocity is only observable in the domain's upper half, and all optimizers assume a zero-initialization in the unobservable bottom half.


\paragraph{Networks.}
The surrogate network approximates the final state $u(x,y\geq 50,t=56)$ in the upper half of the domain from the initial state $u(x,y,t=0)$ and the parameters $x_0, \vec v_0$.
As before, we implement this as G2G (section~\ref{sec:app:baselines-nets}) with five input and two output feature maps, totaling 38.290 parameters.
The G2S JPO network comprises four convolutional layers with 16, 32, 32, and 32 feature maps, respectively, followed by two fully-connected layers with 64 neurons each, resulting in 44.723 total parameters.
Both networks are trained using Adam with a learning rate of $\eta = 0.001$.
%
Fig.~\ref{fig:app:fluid-weight-change} shows the absolute the change in the JPO network weights that results from training.

\paragraph{Additional results.}
As noted in the main text, the high loss value of the JPO is largely due to a fraction of examples with considerably higher loss than the average.
A summary of the individual loss values for $n = 4, 128$ is given in Figs.~\ref{fig:app:fluid-loss-stat-4} and \ref{fig:app:fluid-loss-stat-128}, respectively.
While the neural adjoint method produces the highest loss values before refinement, these all get mapped to relatively small values during the refinement stage.
Meanwhile, the JPO training finds better solutions without refinement since it uses feedback from $F$. However, that also means that the secondary BFGS optimization cannot improve the estimates by nearly as much since many are already close to a (local) minimum.
This leaves a fraction of examples stranded on sub-optimal solutions that contribute significantly to the total loss, despite most problems finding better solutions than BFGS.

Fig.~\ref{fig:app:fluid-improvement} shows the resulting loss and improvement over BFGS, both before and after refinement.
The learning curves for four data set sizes $n$ are shown in Fig.~\ref{fig:app:fluid-curves},
and the parameter evolution of four examples during optimization are shown in Fig.~\ref{fig:app:fluid-params}.

The neural adjoint method nearly always converges to solutions within the training set parameter space, not relying on the boundary loss.
With solution refinement, this results in a mean loss that seems largely independent of $N$ and is slightly lower than the results from direct BFGS optimization.
However, most of this improvement comes from the secondary refinement stage which runs BFGS on the true $F$.
Without solutions refinement, the neural adjoint method yields inaccurate results, losing to BFGS in 98.2\% of cases.

Supervised training does not suffer from examples getting stuck in a local minimum early on.
The highest-loss solutions, which contribute the most to $L$, are about an order of magnitude better than the worst BFGS solutions, leading to a much smaller total loss for $N \geq 16$.
With solution refinement, 64\%, 73\% and 72\% of examples yield a better solution than BFGS for $N = 16, 64, 128$, respectively.

\clearpage
\subsection{Additional experiments} \label{app:additional-experiments}

\begin{figure}[hbt]
    \centering
    \includegraphics[width=\textwidth]{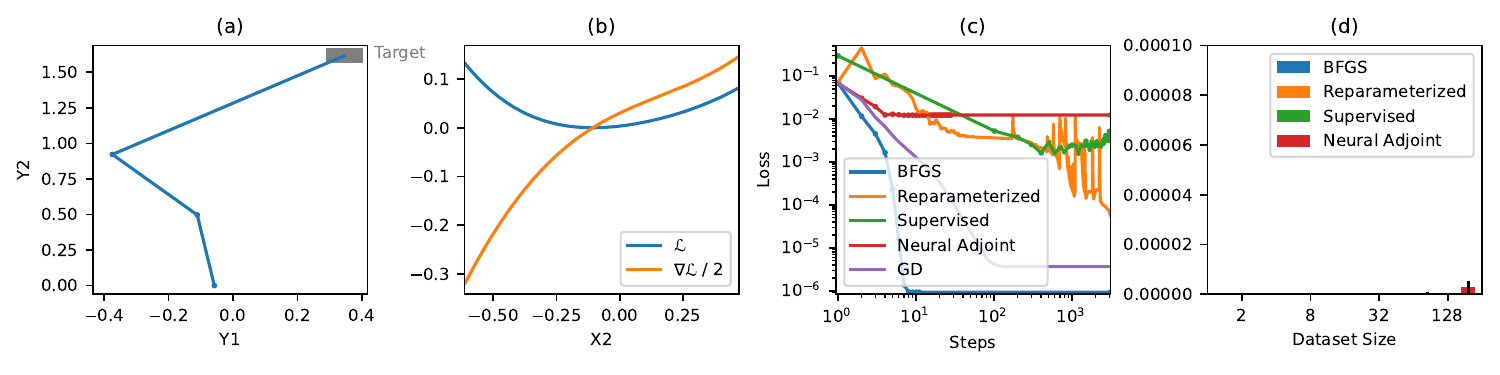}
    \caption{Robotic arm experiment from [RPM20].
    \textbf{(a)}~Task: the arm must be positioned and the joints rotated to reach the target, \textbf{(b)}~corresponding loss and gradient landscape for the first joint angle ($x_2$), \textbf{(c)}~optimization curves without refinement, \textbf{(d)}~refined loss $L/n$ by number of examples $n$, mean and standard deviation over multiple network initializations and data sets. All solutions converge to the global optimum with zero loss.}
    \label{fig:rototic-arm}
\end{figure}

To compare our method with previous work, we replicate the robotic arm experiment from the neural adjoint paper~\citep{NeuralAdjoint}.
However, the original paper did not use the functional form of the forward process, i.e. the true simulation.
When using it, the problems presented there become much easier to solve.
The results are shown in Fig.~\ref{fig:rototic-arm}.
BFGS and gradient descent (GD) manage to reach machine precision accuracy within a couple of iterations while the network approaches take longer to fit the data (3c). The refinement stage then optimizes all examples to machine precision accuracy (3d). While JPO successfully solves these experiments, there is no need to use it since they can be solved perfectly with classical optimizers.
This stands in contrast to the inverse problems shown in the main text which exhibit non-trivial features, such as local optima, zero-gradient regions, or chaotic behavior.
However, this experiment shows, that JPO can be applied to convex optimization problems in the same way.

\end{document}